\pdfoutput=1

\documentclass[11pt]{article}

\usepackage{acl}

\usepackage{times}
\usepackage{latexsym}

\usepackage[T1]{fontenc}

\usepackage[utf8]{inputenc}

\usepackage{microtype}

\usepackage{inconsolata}

\usepackage{graphicx}
\usepackage{subcaption}
\usepackage{amsmath}
\usepackage{booktabs}
\usepackage{makecell}
\usepackage{amssymb}
\usepackage{pifont}
\usepackage{multirow}
\newcommand{\xmark}{\ding{55}}

\usepackage{caption}
\title{\raisebox{-0.5em}{\includegraphics[height=1.8em]{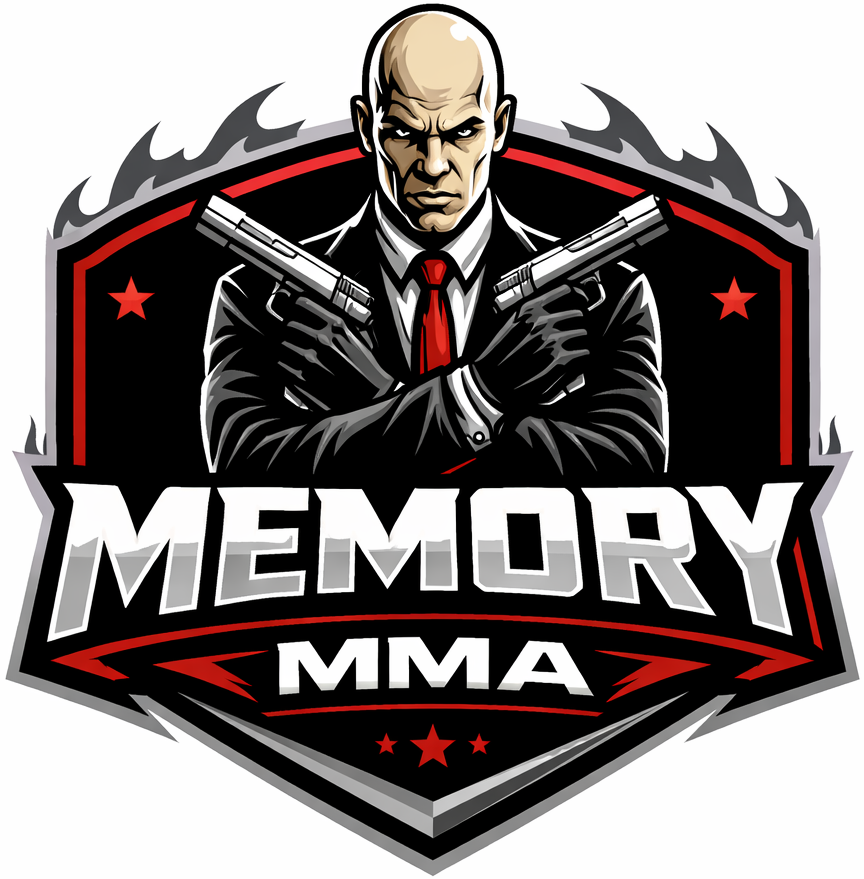}}~MMA: Multimodal Memory Agent}

\author{Yihao Lu$^{*}$\quad
Wanru Cheng$^{*}$\quad
Zeyu Zhang$^{*\dag}$\quad
\textbf{Hao Tang}$^{\ddag}$\\
\vspace{0.2cm}
School of Computer Science, Peking University\\
\small $^*$Equal contribution. $^\dag$Project lead. $^\ddag$Corresponding author: bjdxtanghao@gmail.com.}

\begin{document}
\maketitle
\begin{abstract}
Long-horizon multimodal agents depend on external memory; however, similarity-based retrieval often surfaces stale, low-credibility, or conflicting items, which can trigger overconfident errors. We propose \textbf{Multimodal Memory Agent (MMA)}, which assigns each retrieved memory item a dynamic reliability score by combining \emph{source credibility}, \emph{temporal decay}, and \emph{conflict-aware network consensus}, and uses this signal to reweight evidence and abstain when support is insufficient. We also introduce \textbf{MMA-Bench}, a programmatically generated benchmark for belief dynamics with controlled speaker reliability and structured text–vision contradictions. Using this framework, we uncover the ``Visual Placebo Effect'', revealing how RAG-based agents inherit latent visual biases from foundation models. On FEVER, MMA matches baseline accuracy while reducing variance by 35.2\% and improving selective utility; on LoCoMo, a safety-oriented configuration improves actionable accuracy and reduces wrong answers; on MMA-Bench, MMA reaches 41.18\% Type-B accuracy in Vision mode, while the baseline collapses to 0.0\% under the same protocol.
Code is available at~\url{https://github.com/AIGeeksGroup/MMA}.
\end{abstract}

\section{Introduction}

Memory-augmented LLM agents increasingly underpin long-horizon interactive systems that must preserve and update user-specific context over time~\citep{park2023generativeagentsinteractivesimulacra,guo2024largelanguagemodelbased}. Recent memory architectures introduce more structured memory management and control mechanisms, achieving strong results on conversational benchmarks~\citep{wang2025mirix,packer2023memgpt}. Yet, reliability remains a bottleneck when agents must operate under noisy inputs, stale information, and mutually inconsistent memories.

\begin{figure}[t]
    \centering
    \includegraphics[width=1\linewidth]{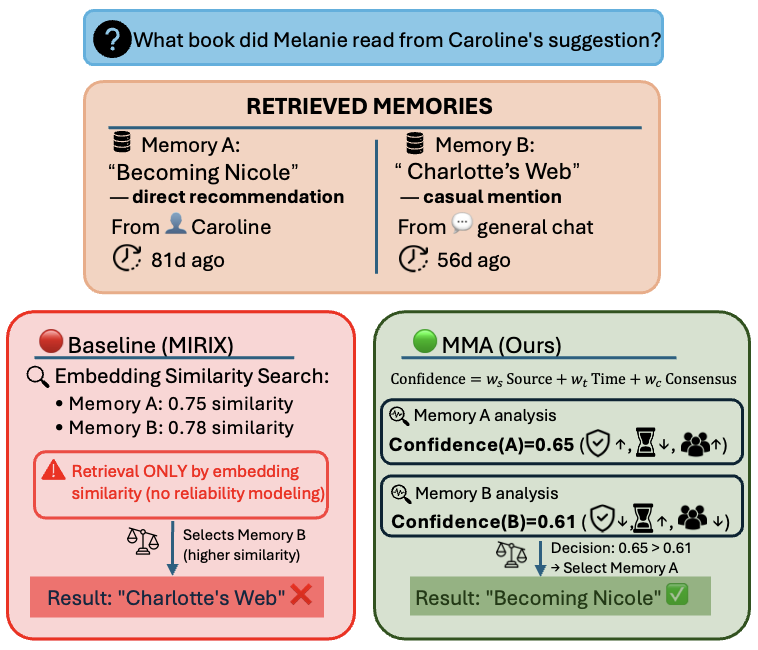}
    \caption{\textbf{Retrieval Trap Case Study.} MIRIX fails by retrieving the high-similarity but irrelevant Memory B. MMA correctly identifies the credible Memory A using multi-dimensional reliability signals.
}
    \label{fig:case_study}
    \vspace{-0.4cm}
\end{figure}

A core limitation is that many memory systems implicitly treat retrieved items as equally reliable by default during reasoning. In practice, information quality varies substantially: sources differ in credibility, facts become outdated, and new retrievals can contradict previously stored content. Without explicit reliability modeling, low-quality memories can propagate through multi-step inference and amplify downstream errors~\citep{xiong2025memorymanagement}. Compounding this, LLM-based agents can produce fluent but unfaithful outputs (hallucinations) that obscure uncertainty and lead to overconfident responses, raising practical safety risks in real-world use~\citep{Ji_2023}. They often respond even when support is insufficient or inconsistent, producing confident answers that later prove to be incorrect. In safety-critical applications, where mistakes impose real costs, this failure to assess evidential sufficiency and arbitrate conflicts becomes particularly problematic.

Given these challenges, our motivation is twofold: (i) memory-level reliability assessment and (ii) evaluation that is incentive-aligned with epistemic prudence. For unreliable memory propagation, agents need mechanisms that separate trustworthy information from questionable content by accounting for source credibility, temporal recency, and coherence with related memories. For epistemic awareness, agents must detect insufficient evidence and abstain when appropriate~\citep{varshney2023stitchtimesavesnine,kuhn2023semanticuncertainty}. Testing this ability requires incentive-aligned frameworks (e.g., abstention-aware scoring) that credit justified abstention and penalize overconfident mistakes, going beyond accuracy-only metrics~\citep{quach2024conformallm,yadkori2024conformalabstention}. This approach better matches real deployment needs~\citep{geifman2017selectiveclassificationdeepneural}, where admitting uncertainty often beats giving wrong answers with confidence.

To address these challenges, we propose MMA (Multimodal Memory Agent), a confidence-aware memory agent with selective prediction capabilities. Our work makes three main contributions. First, we build an inference-time confidence scoring framework at the memory-item level that reweights retrieved memories using source credibility, temporal decay, and conflict-aware network consensus. As shown in Figure~\ref{fig:case_study}, this reliability signal mitigates similarity-based retrieval traps by prioritizing source-credible evidence and discounting stale or weakly supported mentions. Second, we introduce MMA-Bench, a programmatically generated and parameterized benchmark that stresses long-horizon belief revision under controlled source reliability priors and structured text–vision conflicts, with scoring that rewards calibrated abstention and penalizes overconfident errors. Third, we evaluate MMA on FEVER, LoCoMo, and MMA-Bench. On FEVER~\citep{thorne-etal-2018-fever} (500 samples, 3 seeds), MMA matches the MIRIX baseline raw accuracy (59.93\% vs.\ 59.87\%) while reducing standard deviation across seeds by 35.2\% ($\pm$1.62\% vs.\ $\pm$2.50\%), and yields a higher selective score (abstention-aware utility) under abstention reward ($\alpha{=}0.2$: 0.6484 vs.\ 0.6468). On LoCoMo~\citep{maharana2024locomo}, a safety-oriented MMA configuration (without consensus) improves actionable accuracy (79.64\% vs.\ 78.96\%) while reducing wrong answers (298 vs.\ 317). On MMA-Bench, which is deliberately high-noise and retrieval-challenging, MMA achieves 41.18\% Type-B accuracy (reliability inversion) in Vision mode, while the MIRIX baseline records 0.0\% under the same evaluation protocol.

In summary, this work makes three contributions:
\begin{itemize}
    \item We propose the Multimodal Memory Agent (MMA), a dynamic confidence scoring framework that assesses memory reliability through source credibility, temporal decay, and cross-memory consistency.
    \item We introduce MMA-Bench, a diagnostic benchmark that operationalizes belief dynamics under multimodal conflict and controlled reliability priors. Through extensive evaluation, we diagnose the ``Visual Placebo Effect,'' where ambiguous visual inputs can induce unwarranted certainty in RAG-based agents.
    \item We demonstrate improved reliability under risk-aware evaluation across FEVER, LoCoMo, and MMA-Bench, including 35.2\% lower accuracy standard deviation on FEVER, fewer wrong answers on LoCoMo, and 41.18\% Type-B accuracy on MMA-Bench (Vision mode) under the same evaluation protocol.
\end{itemize}

\section{Related Work}

\begin{table*}[t]
\centering
\small
\setlength{\tabcolsep}{2pt}
\resizebox{\textwidth}{!}{%
\begin{tabular}{lcccccc}
\toprule
Benchmark & Setting & Modality & Temp. structure & Paired T--V evidence & Src prior & Epistemic scoring \\
\midrule
LongBench \citep{bai2024longbench} & static LC & Text & static & \xmark & \xmark & accuracy \\
RULER \citep{hsieh2024ruler}       & synth LC  & Text & static & \xmark & \xmark & accuracy \\
LoCoMo \citep{maharana2024locomo}  & LT dialog & Text & multi-session / months & \xmark & \xmark & accuracy \\
FEVER \citep{thorne-etal-2018-fever} & verif.  & Text & static & \xmark & \xmark & accuracy (NEI) \\
\midrule
MMA-Bench (Ours) & LT dialog & Multi & 10 / $\sim$6mo & \checkmark & \checkmark & CoRe \\
\bottomrule
\end{tabular}}
\caption{\textbf{Comparison of Benchmarks Related to Long-horizon Evidence Use.} MMA-Bench complements prior suites by explicitly controlling source reliability priors and pairing multimodal evidence to enable a controlled diagnosis of belief dynamics and epistemic behavior under conflict.}
\label{tab:mma_bench_comparison}
    \vspace{-0.4cm}
\end{table*}

\noindent{\textbf{Memory-Augmented LLM Agents.}}
Memory-augmented agents extend long-horizon interaction by writing to external memory and retrieving relevant items at inference time \citep{packer2023memgpt,wang2025mirix}. Research improves this retrieval-and-inject pipeline through structured/typed memory with specialized modules \citep{wang2025mirix}, context-budgeted memory management with paging and hierarchies \citep{packer2023memgpt,kang2025memoryos,li2025memos}, and lifecycle operations such as versioning and conflict handling \citep{li2025memos}. Other approaches compress or synthesize memory representations to reduce long-horizon overhead \citep{zhou2025mem1,zhang2025memgen} or organize memories into evolving networks for indexing and updates \citep{xu2025amem}. At the same time, empirical evidence suggests that memory policies can induce \emph{experience-following}, where retrieval noise compounds over time \citep{xiong2025memorymanagement}. This points to a complementary gap: most agents still treat retrieved items as uniformly trustworthy despite staleness, low credibility, or inconsistency. MMA operationalizes memory-level reliability with per-item confidence scores that are used directly during downstream reasoning.

\noindent{\textbf{Confidence and Epistemic Mechanisms.}}
Uncertainty estimation and calibration are widely used to mitigate hallucinations. Semantic uncertainty captures meaning-level variability across generations \citep{kuhn2023semanticuncertainty}, and self-consistency methods such as SelfCheckGPT exploit cross-sample disagreement \citep{manakul-etal-2023-selfcheckgpt}. These signals motivate \emph{selective prediction}, including conformal language modeling \citep{quach2024conformallm} and conformal abstention \citep{yadkori2024conformalabstention}; related analyses argue that standard training and evaluation can incentivize systematic overconfidence \citep{kalai2025whyllmshallucinate}. Recent work also explores explicit self-reporting (``confessions'') for monitoring and intervention \citep{joglekar2025trainingllmshonestyconfessions}. Most prior techniques act at the token or response level; in contrast, we target a memory-agent failure mode where \emph{unreliable retrieved memories} trigger overconfident commitments. We evaluate with incentive-aligned scoring that rewards calibrated abstention even when correctness is ambiguous.

\noindent{\textbf{Benchmarks for Multimodal Belief Dynamics.}}
Long-context benchmarks primarily score correctness under extended inputs (LongBench \citep{bai2024longbench}; RULER \citep{hsieh2024ruler}). However, they rarely stress-test belief revision when evidence quality drifts over time, modalities disagree, and agents must decide whether to commit, hedge, or defer. Memory-centric benchmarks move closer to interactive evidence use (LoCoMo \citep{maharana2024locomo}; FEVER \citep{thorne-etal-2018-fever}), but they do not jointly control source reliability priors, temporally evolving multi-session evidence, and structured cross-modal contradictions under abstention-aware evaluation. Recent work highlights multimodal conflict mechanisms \citep{zhang2025modalitiesconflictunimodalreasoning}; we adopt a similar diagnostic lens in long-horizon \emph{memory agents} and focus on how reliability and conflict interact over time. MMA-Bench (Table~\ref{tab:mma_bench_comparison}) fills this gap with controlled priors, paired text–vision evidence, and CoRe (Confidence-and-Reserve) scoring for fine-grained diagnosis of epistemic failures.

Extended discussion of related work is provided in Appendix~\ref{app:related_work_extended}.

\section{The Proposed Method And Benchmark}

\begin{figure*}[t]
\centering
\includegraphics[width=\textwidth]{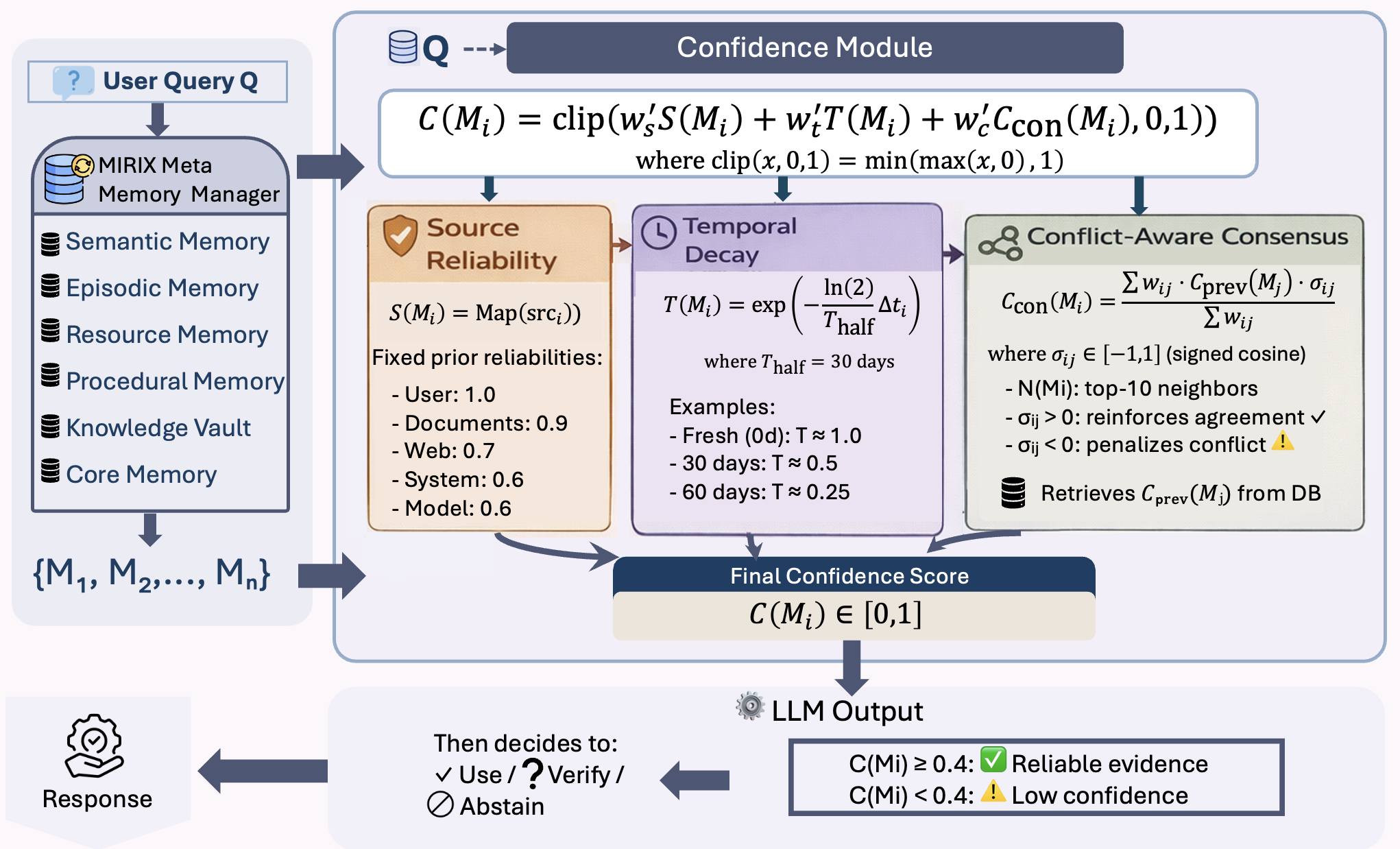}
\caption{\textbf{MMA Framework.} The Confidence Module reweights retrieval via source reliability, temporal decay, and network consensus to modulate reasoning and abstention.}
\label{fig:mma_architecture}
    \vspace{-0.4cm}
\end{figure*}

\subsection{Overview}

We present two contributions: (1) MMA, an agent architecture extending MIRIX \citep{wang2025mirix} with a confidence module for epistemic prudence; and (2) MMA-Bench, a benchmark simulating dynamic social environments to evaluate belief dynamics and calibration under conflict.

\begin{figure*}[t]
    \centering
    \includegraphics[width=\textwidth]{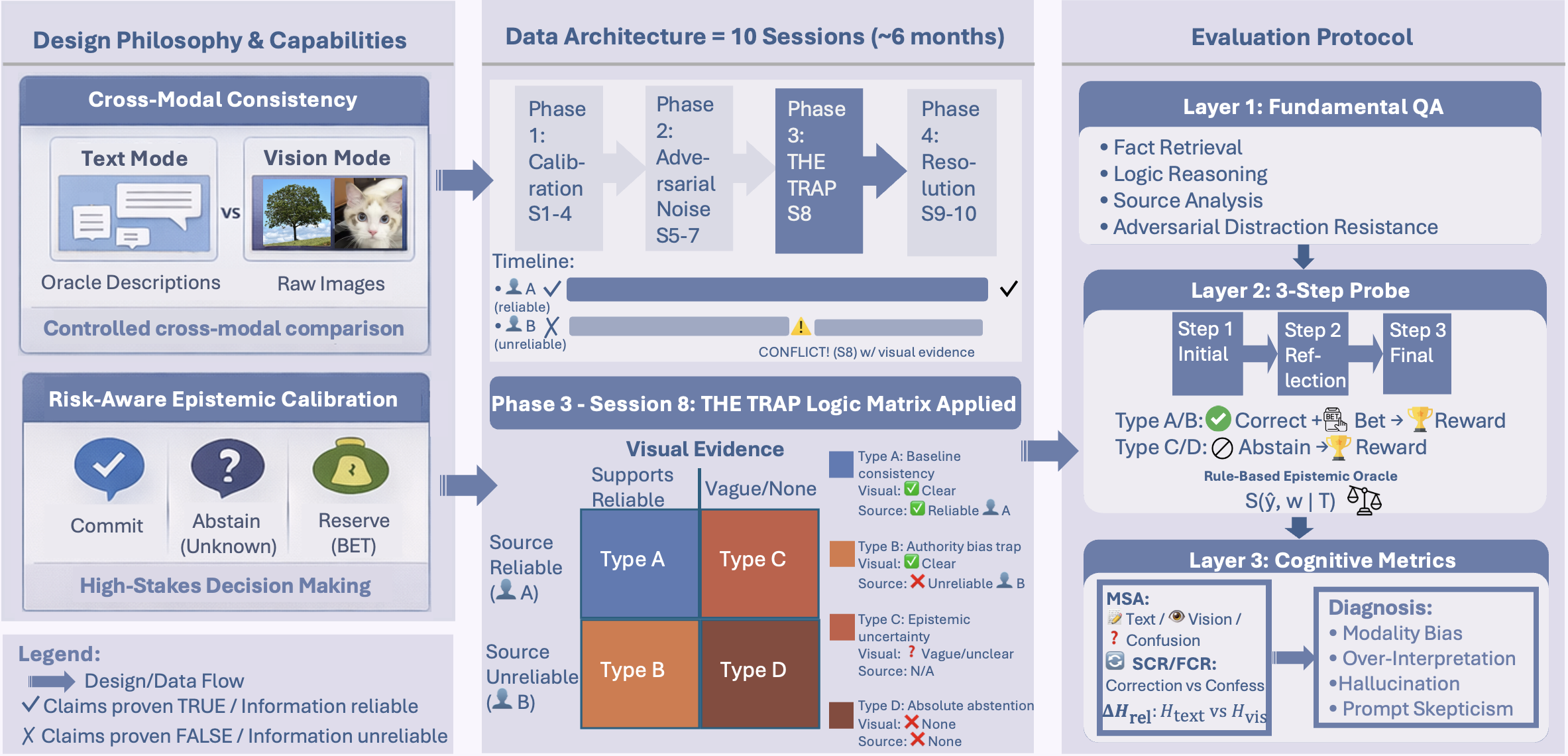}
    \caption{\textbf{MMA-Bench evaluation framework.} The benchmark integrates cross-modal consistency analysis, risk-aware betting, and a $2\times2$ logic matrix for trust conflicts. Performance is assessed through fundamental QA and a 3-step belief probe.}
    \label{fig:framework}
        \vspace{-0.4cm}
\end{figure*}

\subsection{Multimodal Memory Agent (MMA)}

Our approach augments the MIRIX framework with a meta-cognitive reliability layer. Formally, let $\mathcal{M} = \{M_1, M_2, ..., M_N\}$ be retrieved memories for query $Q$. The Confidence Module computes a scalar score $\mathcal{C}(M_i) \in [0, 1]$ to modulate retrieval: high-confidence items are prioritized, while low-confidence items are flagged for potential abstention.

\noindent{\textbf{Confidence Formulation.}}
The confidence score $\mathcal{C}(M_i)$ is a self-normalizing weighted sum of Source ($S$), Time ($T$), and Consensus ($C_{\text{con}}$) components. Using normalized weights $w'_k$, the final score is:
{
\small
\begin{equation}
\mathcal{C}(M_i) = \left[ w'_s S(M_i) + w'_t T(M_i) + w'_c C_{\text{con}}(M_i) \right]_0^1.
\end{equation}
}

\textbf{(1) Source Reliability $S(M_i)$:} We map the memory origin $\text{src}_i$ to a predefined trustworthiness prior. This static score ensures high-quality sources are prioritized:
\begin{equation}
S(M_i) = \text{Map}(\text{src}_i).
\end{equation}

\textbf{(2) Temporal Decay $T(M_i)$:} Models information aging using an exponential decay with a half-life $T_{\text{half}}$:
\begin{equation}
T(M_i) = \exp\left(-\frac{\ln(2)}{T_{\text{half}}} \Delta t_i\right).
\end{equation}

\textbf{(3) Network Consensus $C_{\text{con}}(M_i)$:} 
This metric measures semantic support within the retrieved neighborhood $\mathcal{N}(M_i)$. It acts as a consistency filter, computed as:
\begin{align}
C_{\text{con}}(M_i) &= \frac{\sum_{M_j \in \mathcal{N}(M_i)} w_{ij} \cdot \mathcal{C}(M_j) \cdot \sigma_{ij}}{\sum_{M_j \in \mathcal{N}(M_i)} w_{ij}}, \\
\sigma_{ij} &= \text{sim}_{\cos}(\mathbf{v}_i, \mathbf{v}_j) = \frac{\mathbf{v}_i \cdot \mathbf{v}_j}{\|\mathbf{v}_i\| \|\mathbf{v}_j\|},
\end{align}
where $\sigma_{ij} \in [-1, 1]$ is the Support Factor. Positive values reinforce confidence via alignment, while negative values penalize contradictions.

\subsection{MMA-Bench}
Existing benchmarks for long-context agents predominantly focus on information retrieval or static memory consistency. However, real-world deployment requires agents to navigate conflicting information streams, weigh source reliability against multimodal evidence, and demonstrate epistemic prudence. To address this, we introduce MMA-Bench, a multi-modal benchmark designed to evaluate belief dynamics and cognitive robustness.

\noindent{\textbf{Design Philosophy and Capabilities.}} MMA-Bench evaluates two core dimensions: (1) \textit{Cross-Modal Consistency}, comparing performance in Text Mode (oracle captions) versus Vision Mode (raw images); and (2) \textit{Risk-Aware Epistemic Calibration}, utilizing a betting mechanism to credit justified abstention and penalize overconfidence.

\noindent{\textbf{Data Architecture.}}
Each case is a generated dialogue stream spanning 10 temporal sessions (approx. 6 months). The narrative involves a historically reliable User A and a unreliable User B. The generation pipeline proceeds through four distinct phases: Phase 1 (Calibration, S1-4) implicitly establishes source reliability priors via verifiable events. Phase 2 (Adversarial Noise, S5-7) injects high-volume chit-chat involving entities similar to target facts to rigorously stress-test attention mechanisms. Phase 3 (The Trap, S8) introduces the core multimodal conflict where User B makes a claim supported by visual evidence that contradicts User A. Finally, in Phase 4 (Resolution, S9-10), the ground truth is either resolved or remains unknowable to evaluate abstention capabilities.

\noindent{\textbf{Logic Matrix.}} To systematically evaluate robustness, we formalize a logic matrix that categorizes conflicts into four types based on the interaction between source reliability and visual evidence (Table~\ref{tab:logic_matrix}). This taxonomy is inspired by recent findings on cross-modal inconsistency \citep{zhang2025modalitiesconflictunimodalreasoning}, which highlight that agents often prioritize specific modalities regardless of their reliability.

\begin{table}[h]
\centering
\small
\resizebox{\columnwidth}{!}{%
\begin{tabular}{llp{3.3cm}p{3.8cm}}
\toprule
\textbf{Type} & \textbf{Conflict} & \textbf{Configuration} & \textbf{Target Capability} \\
\midrule
A & Standard & Visuals support reliable User A. & Baseline consistency. \\
B & Inversion & Visuals support unreliable User B. & Overcoming authority bias. \\
C & Ambiguity & Visuals are vague. & Rejecting over-interpretation. \\
D & Unknowable & No valid evidence. & Absolute abstention. \\
\bottomrule
\end{tabular}%
}
\caption{\textbf{Logic Matrix for MMA-Bench.} Categorization of multimodal trust conflicts based on source reliability and visual evidence.}
\label{tab:logic_matrix}
\end{table}

\noindent{\textbf{Evaluation Protocol.}}
We propose a hierarchical framework to dissect performance from basic retrieval to high-level cognitive arbitration.

\textbf{\textit{Layer 1: Fundamental Capabilities.}} This layer assesses foundational skills through standard QA, covering four dimensions: fact retrieval, logic reasoning, source analysis, and adversarial distraction accuracy.

\textbf{\textit{Layer 2: The 3-step Probe \& CoRe Scoring.}}
This layer evaluates the agent's belief state using a 3-step probe, inspired by self-correction mechanisms \citep{joglekar2025trainingllmshonestyconfessions}. To rigorously score calibration, we introduce the \textit{CoRe (Confidence-and-Reserve) Score}, formulated as a rule-based function $S(\hat{y}, \mathbf{w} \mid \mathcal{T})$ conditioned on the logic type~$\mathcal{T}$:
\begin{equation}\small
S = \begin{cases}
\beta \cdot \mathbb{I}(\hat{y}=y^*) + (1-\beta) \cdot \frac{w_{winner}}{100} & \text{if } \mathcal{T} \in \{A, B\} \\
\frac{w_{reserve}}{100} - \gamma \cdot \mathbb{I}(\hat{y} \neq \textsc{Unknown}) & \text{if } \mathcal{T} \in \{C, D\}
\end{cases}
\end{equation}
where $\mathcal{T} \in \{A, B\}$ represents deterministic cases, and $\mathcal{T} \in \{C, D\}$ represents indeterminate cases.

\textbf{\textit{Layer 3: Cognitive Dynamics Metrics.}} To diagnose the mechanics of modality preference and belief revision, we define three metrics. First, \textit{Modality Signal Alignment (MSA)} categorizes the agent's verdict $Y_{model}$ by aligning it with theoretical signal vectors for Text ($S_{text}$) and Vision ($S_{vis}$). In Type B (Inversion), $S_{vis}$ implies \textsc{True} (Trap); in Type C/D, $S_{vis}$ implies \textsc{Unknown} (Uncertainty).
{
\small
\begin{equation}
C(Y_{model}) = \begin{cases} 
\text{Text-Dominant} & \text{if } Y_{model} = S_{text} \\
\text{Vision-Dominant} & \text{if } Y_{model} = S_{vis} \\
\text{Confusion} & \text{otherwise}.
\end{cases}
\end{equation}
}

Second, we quantify the driver of preference using \textit{Relative Reasoning Uncertainty} ($\Delta H_{rel} = 2(H_{text} - H_{vis}) / (H_{text} + H_{vis})$), where a positive value indicates higher certainty in the visual stream. 

Finally, we measure the stability of correct beliefs using the \textit{Self-Correction Rate (SCR)} and the \textit{False Confession Rate (FCR)}. 
The SCR quantifies the probability of correcting an initial error after reflection:
\begin{equation}\small
SCR = \frac{\text{Count}(\text{Step 1} = \text{Wrong} \land \text{Step 3} = \text{Right})}{\text{Count}(\text{Step 1} = \text{Wrong})}.
\end{equation}

Conversely, to diagnose \textit{instructional sycophancy} — the tendency of models to abandon correct beliefs under the pressure of reflection prompts - we define FCR as:
\begin{equation}\small
FCR = \frac{\text{Count}(\text{Step 1} = \text{Right} \land \text{Step 3} = \text{Wrong})}{\text{Count}(\text{Step 1} = \text{Right})}.
\end{equation}

A high FCR relative to SCR indicates that the agent's reasoning is driven by prompt-induced skepticism rather than genuine epistemic calibration.

\section{Experiments}

\subsection{Robustness on Standard Benchmarks}

We first validate MMA on standard text-centric benchmarks to ensure generalizability.

\noindent{\textbf{FEVER (Fact Verification) \citep{thorne-etal-2018-fever}.}} 
As shown in Table \ref{tab:main_results}, MMA matches the baseline's accuracy ($\approx 59.9\%$) but significantly improves stability, reducing the standard deviation by $35.2\%$ ($\pm 1.62\%$ vs. $\pm 2.50\%$). This confirms that our confidence-aware filtering effectively mitigates the stochasticity of retrieval without compromising utility. Full results and component analyzes are detailed in Appendix~\ref{sec:result_fever}.

\begin{table*}[t]
\centering
\small
\begin{tabular}{lccccc}
\toprule
\multirow{2}{*}{\textbf{Method}} & \multicolumn{2}{c}{\textbf{Performance Metrics}} & \multicolumn{2}{c}{\textbf{Prudence Metrics}} & \multirow{2}{*}{\textbf{Stability (Std)}} \\
\cmidrule(lr){2-3} \cmidrule(lr){4-5}
 & \textbf{Raw Acc.} & \textbf{Selective ($\alpha=0.2$)} & \textbf{Abstain Rate} & \textbf{Abstain Prec.} & \\
\midrule
\makecell[l]
{MIRIX (Baseline) \\ \citep{wang2025mirix}} & 59.87\% & 0.6468 & 44.2\% & 45.6\% & $\pm 2.50\%$ \\
\textbf{MMA (Ours)} & \textbf{59.93\%} & \textbf{0.6484} & \textbf{45.3\%} & \textbf{45.8\%} & \textbf{\boldmath$\pm 1.62\%$} \\
\bottomrule
\end{tabular}
\caption{\textbf{Main Results on FEVER.} MMA matches baseline accuracy while significantly reducing performance variance ($\pm$1.62\% vs. $\pm$2.50\%) across seeds.}
\label{tab:main_results}
\vspace{-0.2cm}
\end{table*}

\noindent{\textbf{LoCoMo (Long-Context QA) \citep{maharana2024locomo}.}}
On the sparse LoCoMo benchmark, we observe a density-driven trade-off. While the full consensus module is conservative, the `st' variant (Source + Time) achieves state-of-the-art Utility ($883.6$), outperforming the baseline. This demonstrates the framework's adaptability: consensus is vital for conflict (MMA-Bench) but optional for sparsity. Comprehensive evaluation is provided in Appendix \ref{sec:result_locomo}.

\subsection{Results on MMA-Bench}
\label{subsec:analysis_agent}
\begin{figure*}
    \centering
    \includegraphics[width=1\linewidth]{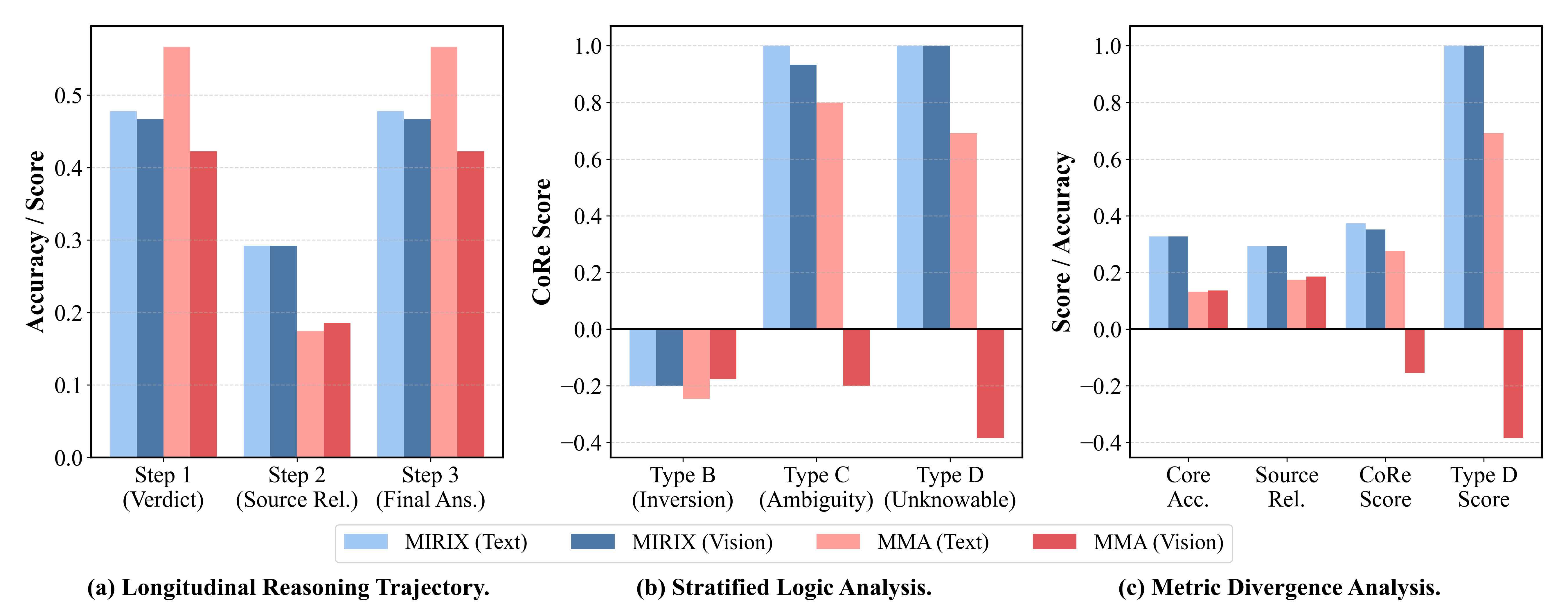}
    \caption{\textbf{Detailed Dynamics Analysis.} (a) Step-wise belief revision; (b) Risk-adjusted scores highlighting visual noise sensitivity; (c) Gap analysis between retrieval accuracy and calibration.}
    \label{fig:cognitive_dynamics_agent}
    \vspace{-0.4cm}
\end{figure*}

We compared the cognitive dynamics of our MMA against the baseline (MIRIX) on the adversarial MMA-Bench. The results, visualized in Figure \ref{fig:cognitive_dynamics_agent}, reveal a fundamental divergence in how confidence-aware agents handle multimodal conflicts compared to standard RAG systems.

\begin{table*}[t]
\centering
\small
\resizebox{\textwidth}{!}{%
\begin{tabular}{llccccc}
\toprule
\multirow{2}{*}{\textbf{Method}} & \multirow{2}{*}{\textbf{Mode}} & \multicolumn{3}{c}{\textbf{Overall Metrics}} & \multicolumn{2}{c}{\textbf{Scenario-Specific Analysis}} \\
\cmidrule(lr){3-5} \cmidrule(lr){6-7}
 & & \textbf{Core Acc.} & \textbf{Verdict Acc.} & \textbf{CoRe Score} & \textbf{Type B Acc.} & \textbf{Type D Score} \\
\midrule
\multirow{2}{*}{\makecell[l] {MIRIX (Baseline) \\ \citep{wang2025mirix}}} & Text & \textbf{30.94\%} & 47.78\% & \textbf{0.37} & 0.00\% & \textbf{1.00} \\
 & Vision & \textbf{32.67\%} & 46.67\% & 0.35 & 0.00\% & \textbf{1.00} \\
\midrule
\multirow{2}{*}{\textbf{MMA (Ours)}} & Text & 13.15\% & \textbf{56.67\%} & 0.28 & 23.53\% & 0.69 \\
 & Vision & 13.55\% & 42.22\% & -0.16 & \textbf{41.18\%} & -0.38 \\
\bottomrule
\end{tabular}%
}
\caption{\textbf{MMA-Bench Main Results.} Comparison across logic types (Type D uses risk-adjusted CoRe scoring). MMA restores agency in Type B conflict and mitigates the visual placebo effect in Type D scenarios.}
\label{tab:agent_performance}
\vspace{-0.4cm}
\end{table*}

\noindent{\textbf{Robustness in Reliability Inversion Scenarios.}} 
In Type B (Reliability Inversion) scenarios, the Baseline exhibits a $100\%$ Confusion rate (defaulting to ``Unknown''). This indicates a failure to engage: due to the high-noise environment, the standard RAG agent fails to retrieve the conflicting evidence required to form a verdict. In contrast, MMA demonstrates active conflict resolution. Despite the difficulty, MMA successfully identifies and prioritizes visual evidence in 41.2\% of cases (Vision Dominant), as visualized in the step-wise verdict distribution (Figure \ref{fig:cognitive_dynamics_agent}). This confirms that the confidence module provides the necessary signal discrimination to attempt resolution, whereas the Baseline remains stagnant due to noise intolerance.

\noindent{\textbf{Qualitative Analysis of Abstention Drivers.}} 
In indeterminate scenarios (Type C and D), the Baseline achieves a deceptively high raw accuracy (Figure \ref{fig:cognitive_dynamics_agent}, Left). However, our analysis suggests that this is an artifact of retrieval limitations rather than epistemic prudence. Qualitative analysis of the response logs reveals that $83.3\%$ of the Baseline's refusals explicitly cite a ``lack of information'', whereas $0\%$ reference source unreliability. This confirms that, due to the high-noise environment, the Baseline simply fails to retrieve the ``trap,'' defaulting to an ``Unknown'' state, which coincidentally aligns with the ground truth. MMA, conversely, actively engages with the noise. In Text Mode, it achieves a high CoRe Score in Type D (Figure \ref{fig:cognitive_dynamics_agent}, Right), demonstrating Intentional Prudence by correctly identifying information gaps based on source reliability analysis.

\noindent{\textbf{Visual Placebo Effect.}} 
We quantify the impact of visual noise by tracking performance shifts in Type D (Unknowable) scenarios. The Baseline (MIRIX) exhibits Zero Visual Sensitivity (Figure \ref{fig:cognitive_dynamics_agent}), maintaining a constant CoRe Score ($\approx1.0$) across modes. This confirms that its apparent stability stems from retrieval blindness—failing to retrieve context makes it immune to visual noise. In stark contrast, MMA suffers a severe regression, with its prudent score collapsing from $0.69$ (Text) to $-0.38$ (Vision). We term this the ``Visual Placebo Effect,'' where the mere presence of visual data bypasses epistemic filters and creates an illusion of evidence.

\subsection{Evolutionary Cognitive Analysis}
\begin{figure}[t]
    \centering
    \includegraphics[width=\columnwidth]{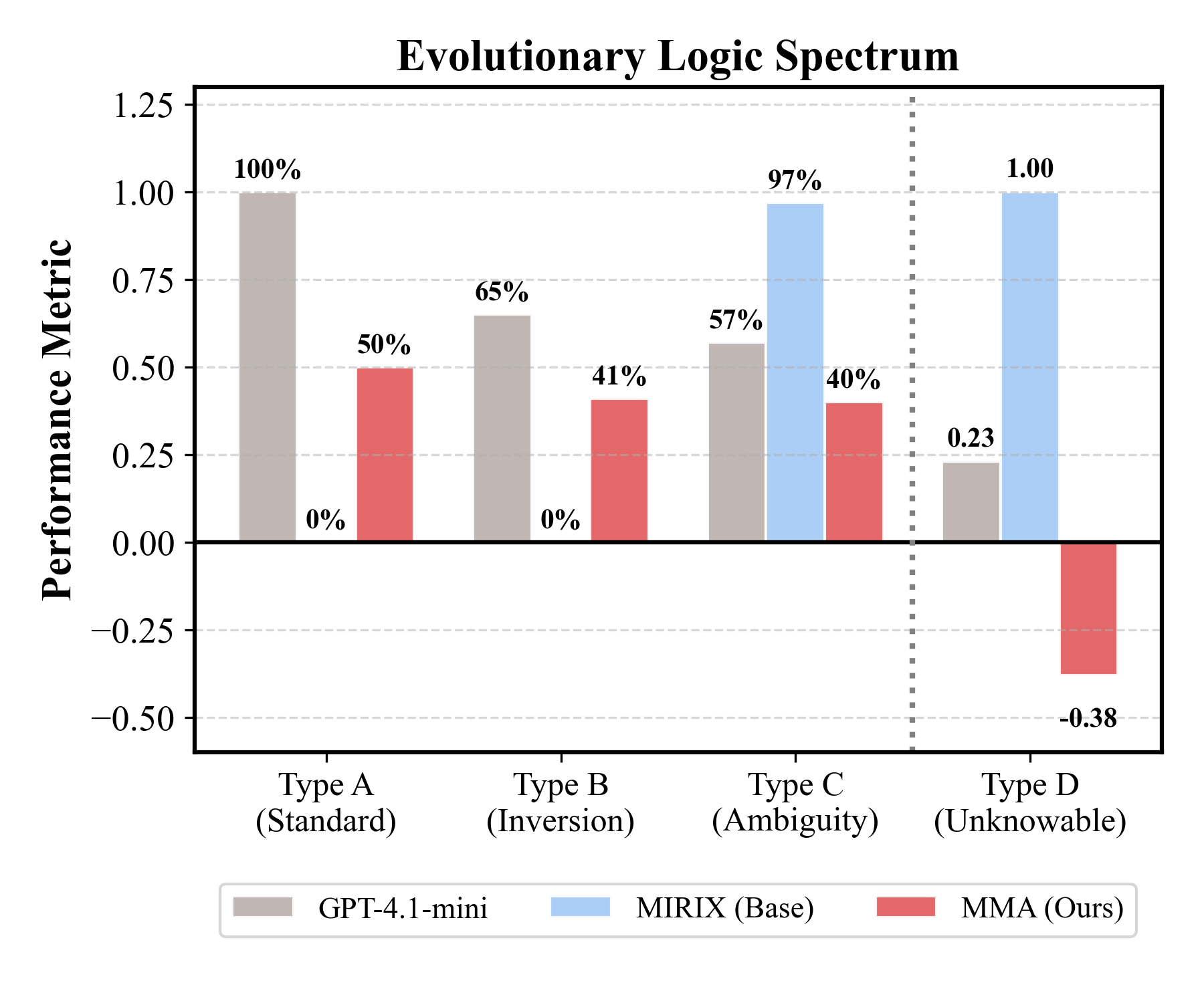}
    \caption{\textbf{Evolutionary Logic Spectrum.} Tracing performance from Foundation Model to MMA. MMA restores agency (Activation) and buffers inherited visual bias (Placebo Effect).}
    \label{fig:evolution_spectrum}
    \vspace{-0.4cm}
\end{figure}

To dissect the mechanics of cognitive enhancement, we analyze the performance trajectory from the foundation model (GPT-4.1-mini, Full Context) to the retrieval-constrained baseline (MIRIX), and finally to our proposed agent (MMA). Figure \ref{fig:evolution_spectrum} visualizes this transformation, illustrating how architectural constraints and confidence modulation interact to shape decision-making behaviors.

\noindent{\textbf{Restoration of Agency in Deterministic Environments.}}
The transition from MIRIX to MMA marks the reactivation of agency. The baseline MIRIX exhibits signs of cognitive paralysis, yielding $0.0\%$ accuracy in Type A and Type B scenarios. Lacking a prior trust distribution, the system is structurally unable to distinguish valid signals from noise, defaulting to inaction. In contrast, MMA functions as a trust catalyst, utilizing Source ($S$) and Time ($T$) modules to restore the capability to form positive verdicts (Type A: $50.0\%$). However, a structural retrieval ceiling persists; neither architecture can replicate the omniscient performance of GPT-4.1-mini ($100\%$ Acc) as the current retrieval implementation restricts them to fragmented evidence (Retrieval Acc $< 35\%$ vs. $80\%$), limiting the upper bound of perception.

\noindent{\textbf{Trade-off Between Ambiguity and Alignment.}}
A critical divergence is observed in Type C (Ambiguity) scenarios. While MIRIX achieves a near-perfect score ($96.7\%$), MMA experiences a significant drop to $40.0\%$. This disparity implies that the success of MIRIX is likely spurious. Response distribution analysis confirms this: in text-based retrieval, $83.3\%$ of the baseline's refusals explicitly cite ``lack of information''. This proves that the baseline defaults to ``Unknown'' due to retrieval blindness rather than intentional epistemic calibration. Conversely, the decline in MMA highlights a side effect of the Consensus Mechanism. In high-entropy environments, enforcing semantic consistency ($C_{\text{con}}$) compels the agent to align with specific signals amidst noise. This suggests that MMA is optimized for active conflict resolution (Type B) at the expense of passivity in ambiguous zones.

\noindent{\textbf{Inheritance of Visual Bias.}}
In Type D (Unknowable) scenarios, we identify a fundamental vulnerability rooted in the foundation model. Quantitative analysis of GPT-4.1-mini reveals a lower entropy for visual signals ($\Delta H_{rel} > 0$), suggesting an inherent tendency to view images as more credible than text. This probabilistic bias is inherited by both MIRIX and MMA. However, its manifestation differs: MIRIX masks this bias through cognitive paralysis, defaulting to ``Unknown'' (Score $1.0$) simply because it fails to engage with the input. MMA, having restored active decision-making, exposes this latent vulnerability. Lacking the global context to correct the inherited bias, MMA is overwhelmed by visual noise (Score $-0.38$). The mere presence of visual data creates an illusion of evidence, leading to high-wager hallucinations.

\noindent{\textbf{Shared Structural Rigidity in Reflection.}}
Our analysis uncovers a systemic dissociation in the self-correction mechanism common to both RAG-based architectures. While GPT-4.1-mini demonstrates high instructional sycophancy (FCR $71.2\%$), both MIRIX and MMA record a numeric FCR of $0\%$. However, this is not due to robustness. A detailed breakdown of the 62 erroneous instances reveals that 100\% (62/62) fall into the ``Logic Collapse'' quadrant: the agents admit error during the reflection step but fail to update the rigid verdict from step 1. This quantitative evidence confirms that the trait of sycophancy is inherited, but the corrective action is mechanically blocked by the architectural rigidity of the pipeline, explaining why both agents acknowledge error during reflection while remaining tethered to their initial erroneous commitment.

\subsection{Ablation Study}
To isolate component contributions, we focus on the critical failure modes exposed on all benchmarks by removing Source ($S$), Time ($T$), and Consensus ($C_{\text{con}}$), as summarized in Table \ref{tab:ablation_mma_data}. Full results are in the Appendices.

\begin{table}[h]
\centering
\small
\resizebox{\columnwidth}{!}{%
\begin{tabular}{lcccc}
\toprule
\multirow{2}{*}{\textbf{Model}} & \multicolumn{2}{c}{\textbf{Deterministic (Type A, Vis)}} & \multicolumn{2}{c}{\textbf{Indeterminate (Type D, Vis)}} \\
\cmidrule(lr){2-3} \cmidrule(lr){4-5}
 & \textbf{Verdict Acc.} & \textbf{Status} & \textbf{CoRe Score} & \textbf{Interpretation} \\
\midrule
\textbf{MMA (Full)} & $\mathbf{50.0\%}$ & \textbf{Robust} & $\mathbf{-0.38}$ & \textbf{Buffered} \\
\midrule
\textbf{tc} (w/o Source) & $0.0\%$ & Paralyzed & $1.00^{\dagger}$ & Artifact of Default \\
\textbf{st} (w/o Consen.) & $36.7\%$ & Unstable & $-0.69$ & Hallucinated \\
\textbf{cs} (w/o Time) & $0.0\%$ & Degraded & $1.00^{\dagger}$ & Artifact of Default \\
\bottomrule
\end{tabular}
}
\caption{\textbf{Ablation results on MMA-Bench (Vision Mode).} $^{\dagger}$Perfect scores in Type D coincide with 0\% accuracy in Type A, indicating system paralysis rather than genuine calibration.}
\label{tab:ablation_mma_data}
\end{table}

\noindent{\textbf{Impact of Source Reliability ($S$).}}
Comparison with the `tc' variant (w/o Source) reveals that source credibility is a prerequisite for agency. On MMA-Bench, removing $S$ leads to Cognitive Paralysis, where the agent yields $0.0\%$ accuracy in deterministic scenarios (Type A/B). This distinct failure pattern proves that without a prior trust distribution, the system is mechanically incapable of distinguishing signal from noise, defaulting to inaction regardless of the benchmark.

\noindent{\textbf{Impact of Network Consensus ($C_{\text{con}}$).}}
The `st' variant (w/o Consensus) highlights the role of consensus as a safety buffer. While `st' performs well in sparse contexts (LoCoMo), it lacks the arbitration logic to handle multimodal noise. In MMA-Bench Type D scenarios, `st' suffers a catastrophic score collapse to $-0.69$, indicating that isolated visual signals easily override textual caution. By reintroducing consensus, MMA buffers this drop to $-0.38$, effectively filtering out hallucinations that lack semantic support from the memory neighborhood.

\noindent{\textbf{Impact of Temporal Decay ($T$).}}
The `cs' variant (w/o Time) demonstrates a critical failure in cross-modal stability. Without temporal decay, historical noise that is manageable in text-only settings becomes overwhelming when compounded by high-dimensional visual features. This is evidenced by the performance evaporation in MMA-Bench Vision Mode ($0.0\%$ Acc in Type A), confirming that temporal awareness is essential for maintaining a viable signal-to-noise ratio in dynamic environments.

\section{Conclusion}

In this work, we introduce \textbf{MMA}, a confidence-aware memory framework transforms passive memory storage into active epistemic filtering. Through systematic evaluation on FEVER, LoCoMo, and our MMA-Bench, we demonstrate that explicit reliability modeling significantly improves stability and calibrated abstention. 

First, we propose a dynamic scoring mechanism that significantly improves stability ($\pm 1.62\%$ vs. $\pm 2.50\%$ on FEVER) and enables calibrated abstention. Second, through our novel \textbf{MMA-Bench}, we identify the ``Visual Placebo Effect'', revealing that multimodal agents inherit a latent visual bias from foundation models. MMA effectively mitigates this bias, restoring decision-making agency in deterministic scenarios where baselines suffer from cognitive paralysis. Third, empirical results demonstrate that MMA achieves a superior risk-coverage trade-off, delivering high utility in safety-critical environments. MMA represents a step toward epistemic prudence in agent design, providing cognitive guardrails for high-stakes applications.

\clearpage
\section*{Limitations}

While MMA enhances reliability, two limitations warrant future exploration. \textbf{First, reliance on upstream retrieval recall:} As a post-retrieval module, MMA can filter out hallucinations but cannot rectify the absence of evidence if the underlying RAG system fails to retrieve relevant context. \textbf{Second, the sparsity-consensus trade-off:} Our analysis on LoCoMo suggests that strict consensus enforcement can be conservative in low-density information environments. Future work could explore adaptive gating mechanisms that dynamically toggle consensus based on context entropy.

\bibliography{custom}

\clearpage
\appendix
\section{Extended Related Work}
\label{app:related_work_extended}
This section expands on related research that is only briefly mentioned in the main paper due to space constraints. We provide additional background on (i) memory architectures and control policies for long-horizon agents, (ii) compressed or synthesized memory representations, (iii) uncertainty and selective-prediction mechanisms, and (iv) benchmark design for long-context and multimodal belief dynamics. These discussions offer supporting context for the reliability- and abstention-focused setting studied in this work.

\paragraph{Memory Architectures And Control Policies.}
Beyond basic retrieval-and-inject, recent systems emphasize \emph{explicit control} over what is written, how it is indexed, and when it is surfaced to the model.
MIRIX proposes typed memory with dedicated modules for writing, retrieval, and routing, enabling finer-grained control over what enters the reasoning context \citep{wang2025mirix}.
MemGPT treats the context window as a managed resource and introduces paging between the prompt and external storage \citep{packer2023memgpt}.
Related ``memory OS'' lines of work propose multi-tier hierarchies and policy-driven memory operations to mitigate context growth and reduce retrieval noise \citep{kang2025memoryos,li2025memos}.
These approaches primarily improve \emph{organization} and \emph{access}, but they typically do not provide an explicit epistemic signal that differentiates trustworthy from questionable retrieved content at the level of individual memory items.

\paragraph{Compressed And Synthesized Memory Representations.}
A complementary direction reduces long-horizon overhead by compressing interaction history or synthesizing latent memory.
MEM1 compresses trajectories into compact states intended to support long-horizon reasoning under a constant-memory interface \citep{zhou2025mem1}.
MemGen generates latent memory conditioned on agent state, aiming to preserve salient information while avoiding unbounded growth \citep{zhang2025memgen}.
A-MEM further organizes memories as evolving note-like networks to support dynamic indexing and updates \citep{xu2025amem}.
While these representations can improve scalability, they do not directly resolve the \emph{reliability} issue when retrieved items are stale, low-credibility, or mutually inconsistent.

\paragraph{Error Accumulation in Long-horizon Memory Agents.}
Recent empirical evidence suggests that memory policies can induce \emph{experience-following}, where retrieval noise compounds across turns and systematically steers future behavior \citep{xiong2025memorymanagement}.
This phenomenon motivates reliability-aware mechanisms that act \emph{before} noisy memories enter downstream reasoning, rather than only mitigating errors at the final response stage.

\paragraph{Uncertainty Signals, Selective Prediction, And Self-reporting.}
Uncertainty estimation for language generation has been studied from multiple angles.
Semantic uncertainty estimates meaning-level variability across alternative generations \citep{kuhn2023semanticuncertainty}.
SelfCheckGPT uses cross-sample disagreement as a black-box signal for hallucination risk \citep{manakul-etal-2023-selfcheckgpt}.
Such signals connect naturally to \emph{selective prediction}, where a model answers only when sufficiently confident:
conformal language modeling provides coverage-style guarantees for language outputs \citep{quach2024conformallm}, and conformal abstention explicitly optimizes the decision to refrain under uncertainty \citep{yadkori2024conformalabstention}.
Complementary analyses argue that conventional training and evaluation can incentivize systematic overconfidence and guessing \citep{kalai2025whyllmshallucinate}.
Recent work also explores explicit self-reporting mechanisms (e.g., ``confessions'') to surface potential mistakes for monitoring and intervention \citep{joglekar2025trainingllmshonestyconfessions}.
Most of these approaches operate at the token or response level; our focus differs in that we attach uncertainty to \emph{retrieved memory items} and use it to modulate reasoning and abstention when retrieval is unreliable.

\paragraph{Benchmarks for Long-context Reasoning And Interactive Memory.}
Long-context benchmarks primarily score correctness under extended inputs.
LongBench provides a multilingual, multi-task suite for long-context understanding \citep{bai2024longbench}, and RULER uses configurable synthetic probes to study effective context use beyond naive retrieval \citep{hsieh2024ruler}.
Memory-centric benchmarks move closer to interactive settings:
LoCoMo evaluates very long-term conversational memory over extended dialogs \citep{maharana2024locomo}, while FEVER evaluates evidence-based verification with a dedicated insufficient-evidence label \citep{thorne-etal-2018-fever}.
However, these suites do not jointly control (i) \emph{source reliability priors}, (ii) \emph{temporally evolving multi-session evidence}, and (iii) \emph{structured cross-modal contradictions} under an abstention-aware utility.

\paragraph{Multimodal Conflict and Modality Preference Dynamics.}
Recent analysis suggests that when modalities conflict, the model's preference can be governed by relative unimodal reasoning uncertainty \citep{zhang2025modalitiesconflictunimodalreasoning}.
We adopt a related diagnostic lens but place it in a long-horizon memory-agent setting where reliability evolves over time and conflicts arise from both source priors and multimodal evidence.
MMA-Bench is designed to isolate these dynamics with paired text–vision evidence, controlled priors, and CoRe scoring, enabling fine-grained diagnosis of epistemic failures beyond accuracy-only metrics.

\section{Results on FEVER Benchmark}
\label{sec:result_fever}

\paragraph{\textbf{Overall Performance and Stability.}}
To rigorously evaluate the effectiveness of our proposed Multimodal Memory Agent (MMA) framework, we conducted experiments on the FEVER benchmark using three random seeds (42, 922, 2025). For fair comparison, we align the evaluation scope to the first 500 samples for both the baseline and MMA across all seeds. Table~\ref{tab:main_results} summarizes the aggregated performance metrics.

Statistical Robustness. While the baseline achieves a comparable raw accuracy average to MMA ($\approx 59.9\%$), it exhibits significant instability across different seeds. Specifically, the baseline's accuracy fluctuates widely with a high standard deviation ($\pm 2.50\%$). In contrast, MMA demonstrates superior robustness, maintaining a significantly lower variance ($\pm 1.62\%$). This indicates that our confidence-aware mechanism effectively mitigates the stochasticity inherent in retrieval-augmented generation.

\paragraph{\textbf{Prudence and Calibration Analysis.}}
A key contribution of our framework is enhancing the agent's ability to ``know what it does not know.'' We analyze this through the lens of abstention behavior and selective scoring.

Improved Precision in Abstention. As shown in the detailed breakdown, MMA adopts a more prudent strategy, abstaining on average $226.3$ times per 500 samples, compared to $221.0$ for the baseline. Crucially, this conservatism is well-calibrated: MMA correctly identifies ``Not Enough Info'' (NEI) cases more frequently than the baseline (Average Correct Abstain: $103.7$ vs. $100.7$). This suggests that MMA is not merely silent, but selectively silent when information is truly insufficient.

\begin{figure}[t]
    \centering
    \begin{subfigure}[t]{0.49\linewidth}
        \centering
        \includegraphics[width=\linewidth]{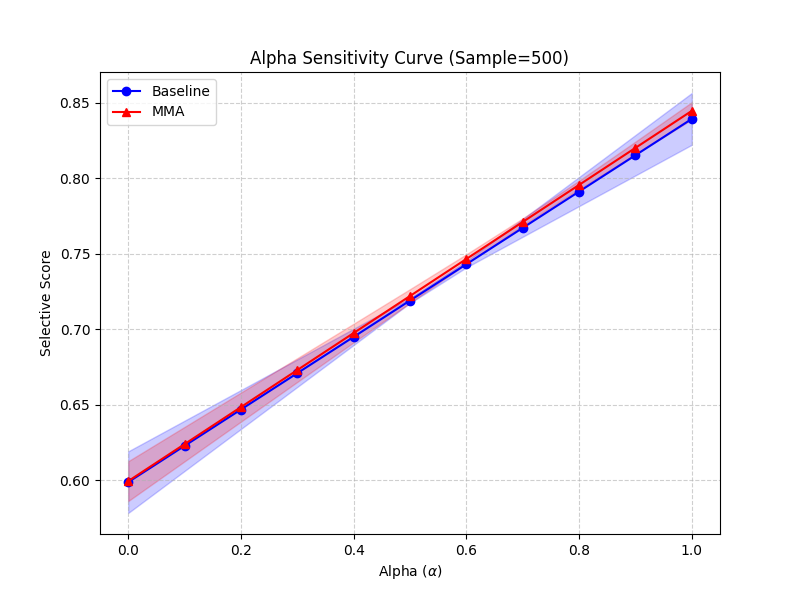}
        \caption{\textbf{Sensitivity Analysis of Abstention Reward ($\alpha$).}}
        \label{fig:alpha_sensitivity_fever}
    \end{subfigure}
    \hfill
    \begin{subfigure}[t]{0.49\linewidth}
        \centering
        \includegraphics[width=\linewidth]{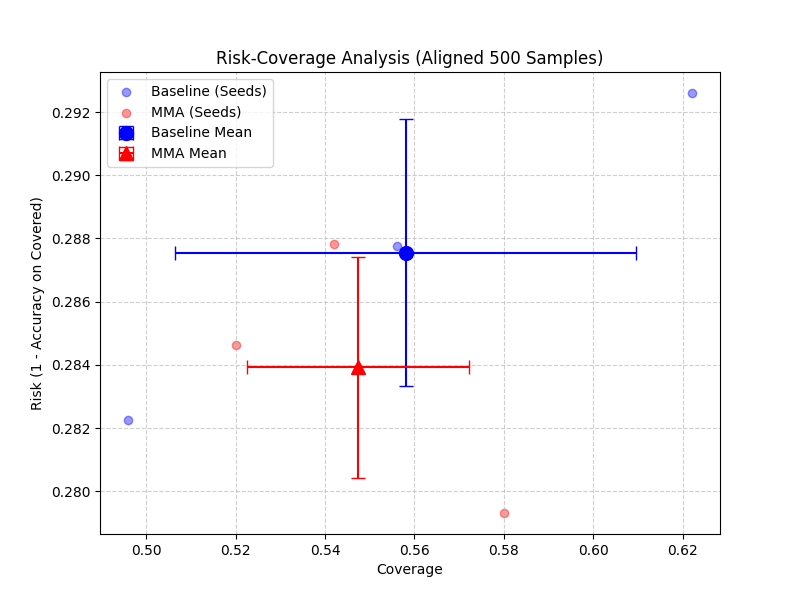}
        \caption{\textbf{Risk-Coverage Analysis.}}
        \label{fig:risk_coverage_fever}
    \end{subfigure}
    \caption{\textbf{Selective Prediction Analysis on FEVER.} MMA consistently outperforms the Baseline under abstention-based risk control, achieving higher utility and lower risk across evaluation settings.}
    \label{fig:selective_prediction_fever}
\end{figure}

Sensitivity to Abstention Reward ($\alpha$). To further quantify the utility of our model in risk-sensitive scenarios, we evaluated the Selective Score with varying abstention reward parameters ($\alpha$). As illustrated in Figure~\ref{fig:alpha_sensitivity_fever}, at the starting point ($\alpha=0$) where no credit is given for abstention, both models exhibit nearly identical raw accuracy ($\approx 59.9\%$). However, as $\alpha$ increases—simulating scenarios where safety is prioritized—the MMA curve (red) consistently rises above the baseline (blue). Notably, the error band for MMA is visibly narrower than that of the baseline, confirming that our method consistently delivers higher utility with lower variance.

Risk-Coverage Trade-off. We further visualize the relationship between the model's willingness to answer (Coverage) and the error rate of those answers (Risk) in Figure~\ref{fig:risk_coverage_fever}. The MMA data points (red) cluster towards the bottom-left quadrant relative to the baseline (blue), indicating lower coverage but simultaneously lower risk. By filtering out low-confidence retrieval results through our consensus mechanism, MMA sacrifices a small portion of coverage to ensure that the provided answers maintain a higher standard of correctness. This trade-off is highly desirable for trusted agents, where hallucinations are costly.

\paragraph{\textbf{Performance on Long-Context Text Benchmarks.}} 
To validate robustness in non-adversarial settings, we also evaluated MMA on the LoCoMo benchmark. Results indicate a distinct trade-off driven by information sparsity: while the Full Model prioritizes prudence (lower coverage), the `st' variant (Source + Time) effectively balances safety and retrieval, achieving the highest Actionable Accuracy ($79.64\%$) and Utility ($883.6$), slightly surpassing the baseline. This demonstrates the framework's adaptability to varying density contexts. Comprehensive results are presented in Appendix \ref{sec:result_locomo}.

\begin{figure*}[t]
    \centering
    \begin{subfigure}[b]{0.32\textwidth}
        \centering
        \includegraphics[width=\textwidth]{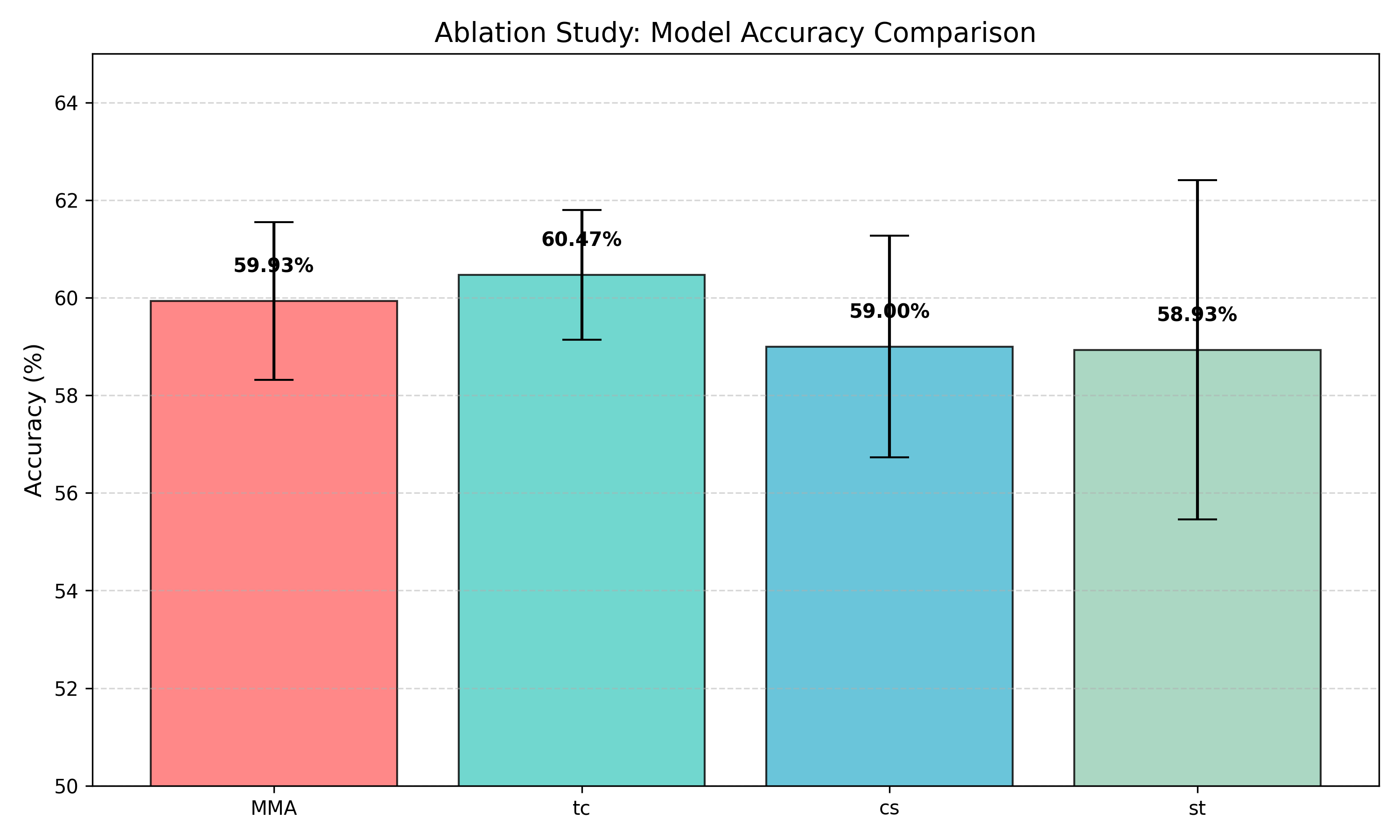} 
        \caption{\textbf{Stability Analysis.}}
        \label{fig:ablation_acc_fever}
    \end{subfigure}
    \hfill
    \begin{subfigure}[b]{0.32\textwidth}
        \centering
        \includegraphics[width=\textwidth]{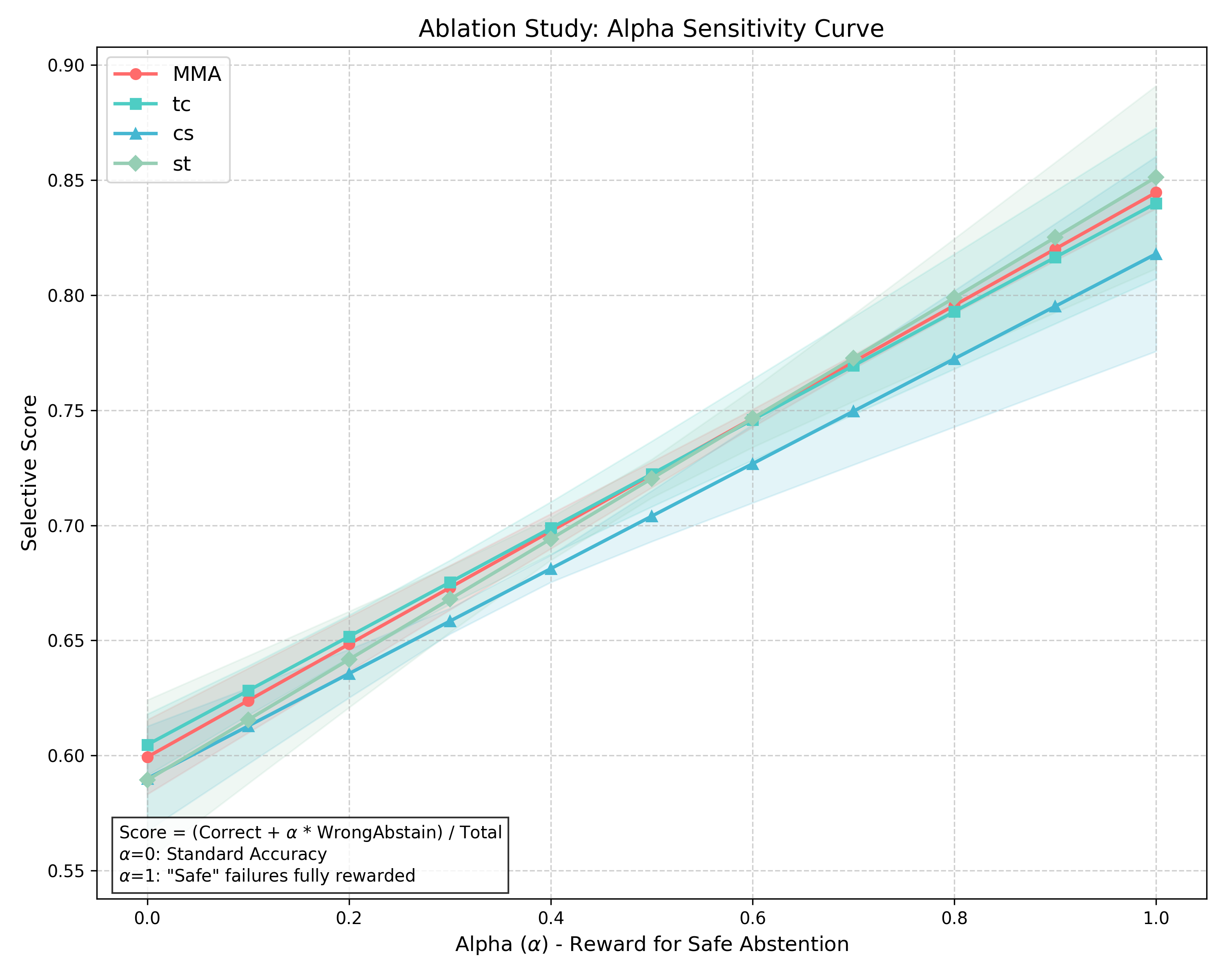}
        \caption{\textbf{Prudence Analysis.}}
        \label{fig:alpha_curve_fever}
    \end{subfigure}
    \hfill
    \begin{subfigure}[b]{0.32\textwidth}
        \centering
        \includegraphics[width=\textwidth]{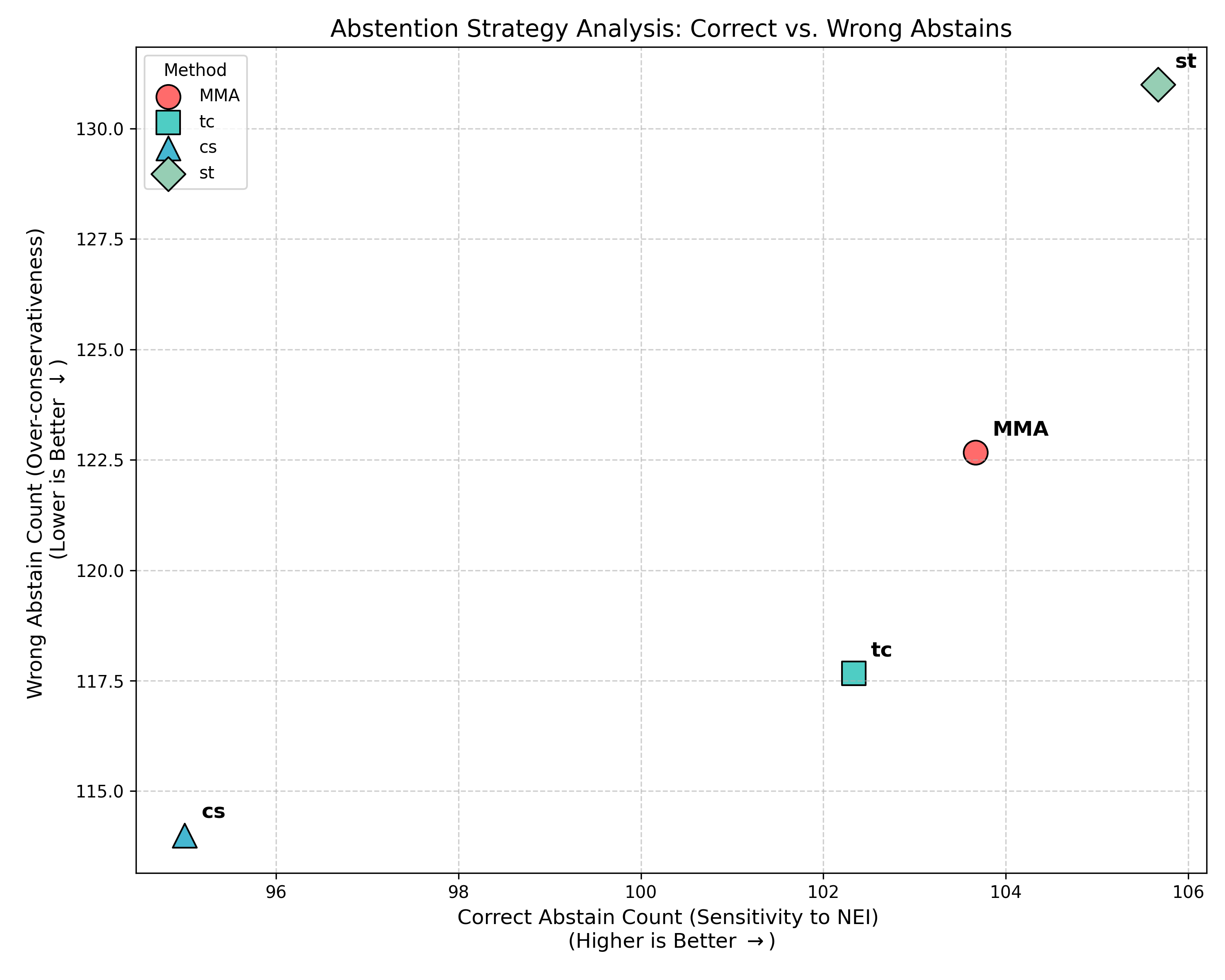} 
        \caption{\textbf{Strategy Analysis.}}
        \label{fig:ablation_scatter_fever}
    \end{subfigure}
    
    \caption{\textbf{Ablation Study Results on FEVER.} We compare the Full Model (MMA) against variants without Consensus (`st'), without Time (`cs'), and without Source (`tc'). (a) Shows that removing Consensus drastically increases variance. (b) Shows that MMA maintains high utility under strict prudence requirements (high $\alpha$). (c) Visualizes the trade-off between sensitivity and conservativeness.}
    \label{fig:ablation_main_fever}
\end{figure*}

\paragraph{\textbf{Ablation Analysis on FEVER.}}
We evaluated the variants on the FEVER dataset ($N=500$) across three random seeds. The comprehensive results are presented in Table \ref{tab:ablation_data_fever} and Figure \ref{fig:ablation_main_fever}.

\begin{table}[h]
\centering
\small
\resizebox{\columnwidth}{!}{%
\begin{tabular}{lccccc}
\toprule
\textbf{Mode} & \textbf{Components} & \textbf{Raw Acc. (\%)} & \textbf{Act. Acc. (\%)} & \textbf{Correct Abstain} & \textbf{Wrong Abstain} \\
\midrule
\textbf{MMA (Ours)} & $S + T + C_{\text{con}}$ & $59.93 \pm 1.62$ & $71.61 \pm \mathbf{0.43}$ & $103.7 \pm 4.7$ & $122.6 \pm 11.2$ \\
\midrule
\textbf{tc} (w/o Source) & $T + C_{\text{con}}$ & $\mathbf{60.47 \pm 1.33}$ & $71.61 \pm 2.54$ & $102.3 \pm 12.6$ & $117.7 \pm 20.6$ \\
\textbf{cs} (w/o Time) & $S + C_{\text{con}}$ & $59.00 \pm 2.27$ & $68.96 \pm 2.35$ & $95.0 \pm 13.2$ & $\mathbf{114.0 \pm 32.1}$ \\
\textbf{st} (w/o Consen.) & $S + T$ & $58.93 \pm 3.48$ & $\mathbf{72.05 \pm 2.34}$ & $\mathbf{105.7 \pm 16.1}$ & $131.0 \pm 36.4$ \\
\bottomrule
\end{tabular}
}
\caption{\textbf{Ablation results on FEVER ($N=500, \text{Seeds}=3$).} \textbf{Act. Acc. (Actionable Accuracy)} denotes the precision of non-abstained responses (Mean $\pm$ Std). Notably, while \textbf{MMA} shares a similar mean accuracy with other variants, it achieves significantly lower variance ($\sigma=0.43$), demonstrating superior stability compared to \textbf{tc} ($\pm 2.54$) and \textbf{st} ($\pm 2.34$).}
\label{tab:ablation_data_fever}
\end{table}

\textbf{Impact of Temporal Decay ($T$): Enabling Prudence.} 
The capability to ``know what you don't know'' is critical for reliable agents. As shown in Table \ref{tab:ablation_data_fever}, the removal of the temporal module (Model `cs') results in the lowest number of \textit{Correct Abstentions} (95.0) and the lowest \textit{Actionable Accuracy} (68.96\%). Figure \ref{fig:alpha_curve_fever} further illustrates that `cs' underperforms significantly as the reward for safe abstention ($\alpha$) increases. This suggests that without temporal awareness, the agent fails to identify outdated information, leading to overconfident hallucinations rather than prudent refusals.

\textbf{Impact of Network Consensus ($C_{\text{con}}$): Ensuring Stability.}
While the `st' variant (w/o Consensus) achieves a high mean Actionable Accuracy ($72.05\%$), it suffers from severe instability ($\sigma \approx 2.34\%$) and excessive conservatism (highest Wrong Abstains: 131.0). In stark contrast, the Full Model (MMA) achieves a comparable Actionable Accuracy ($71.61\%$) but with a remarkably low standard deviation of $\mathbf{\pm 0.43\%}$. As visualized in Figure \ref{fig:ablation_acc_fever}, the inclusion of our conflict-aware consensus mechanism effectively smooths out retrieval noise, ensuring consistent and reproducible behavior across different initializations.

\textbf{Impact of Source Reliability ($S$).}
Interestingly, Mode `tc' (w/o Source) achieves the highest raw accuracy on FEVER. We attribute this to the homogeneity of the FEVER dataset (Wikipedia-based), where source credibility is uniformly high. However, the Source module becomes indispensable in adversarial scenarios with mixed reliability.

The Full Model (MMA) achieves the optimal trade-off. It avoids the ``blind guessing'' of `cs' and the ``erratic conservatism'' of `st', providing a stable, prudent, and trustworthy solution for fact verification.

\section{Results on LoCoMo Benchmark}
\label{sec:result_locomo}

\begin{figure*}[t]
    \centering
    \begin{subfigure}[b]{0.32\textwidth}
        \centering
        \includegraphics[width=\textwidth]{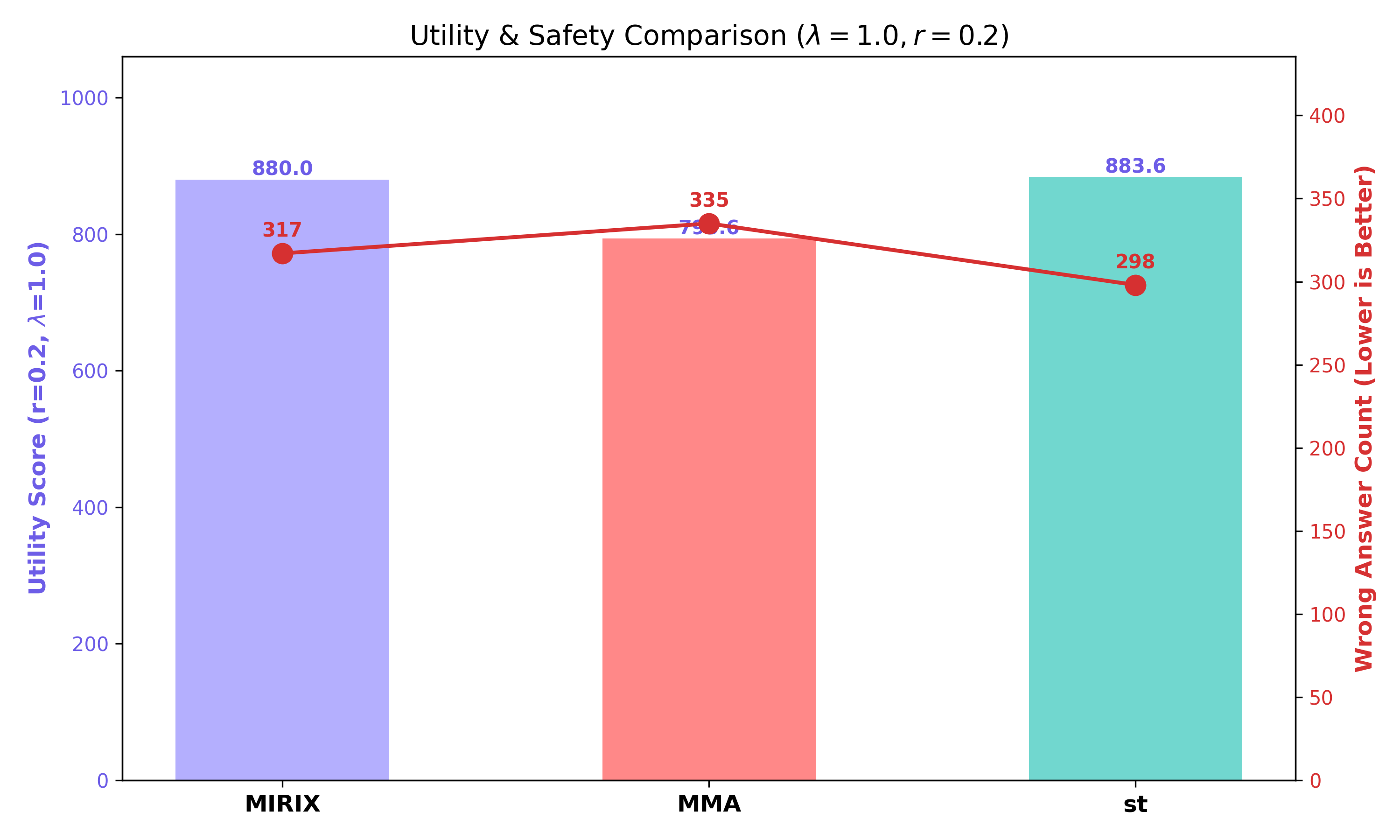} 
        \caption{\textbf{Utility vs. Safety.}}
        \label{fig:locomo_utility}
    \end{subfigure}
    \hfill
    \begin{subfigure}[b]{0.32\textwidth}
        \centering
        \includegraphics[width=\textwidth]{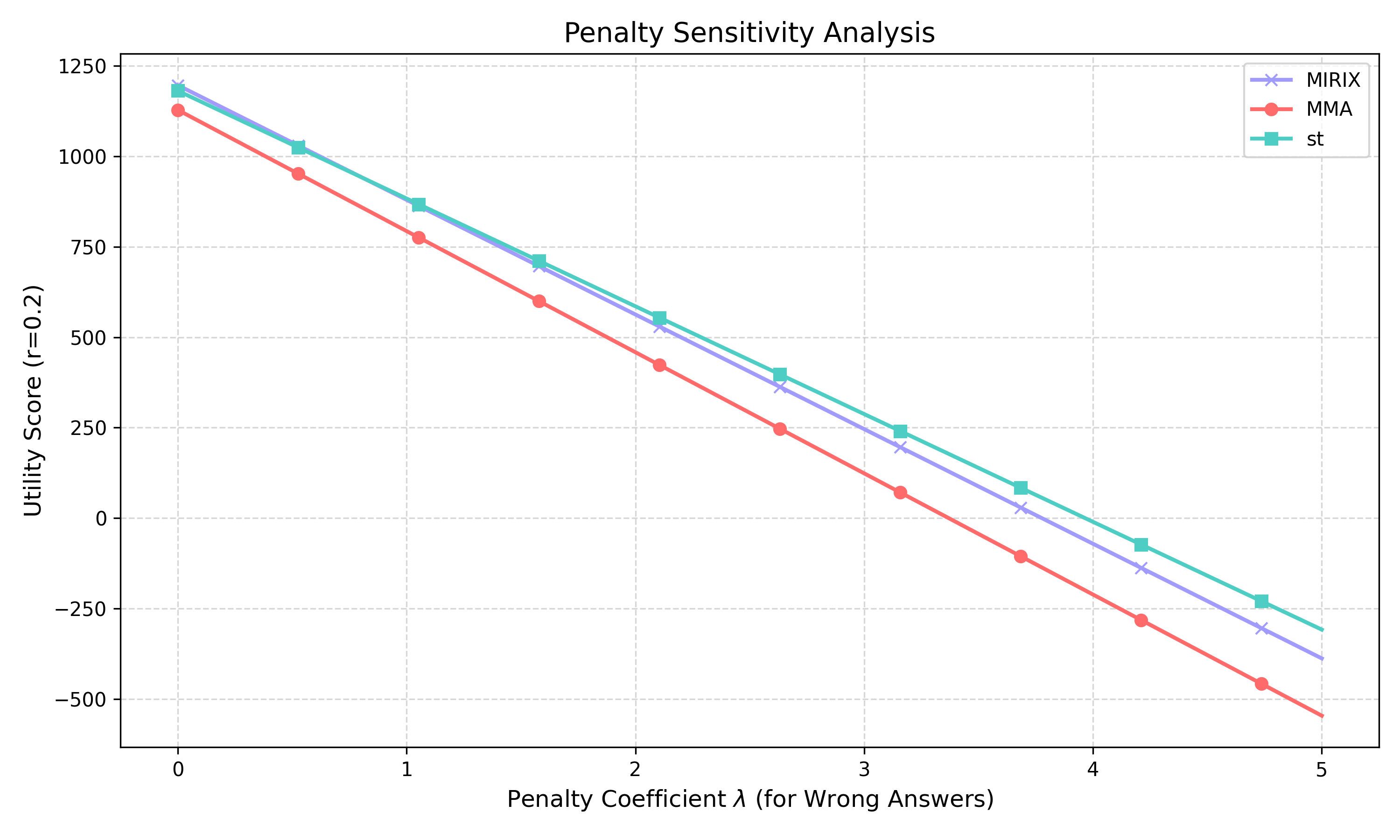}
        \caption{\textbf{Penalty Sensitivity ($\lambda$).}}
        \label{fig:locomo_penalty}
    \end{subfigure}
    \hfill
    \begin{subfigure}[b]{0.32\textwidth}
        \centering
        \includegraphics[width=\textwidth]{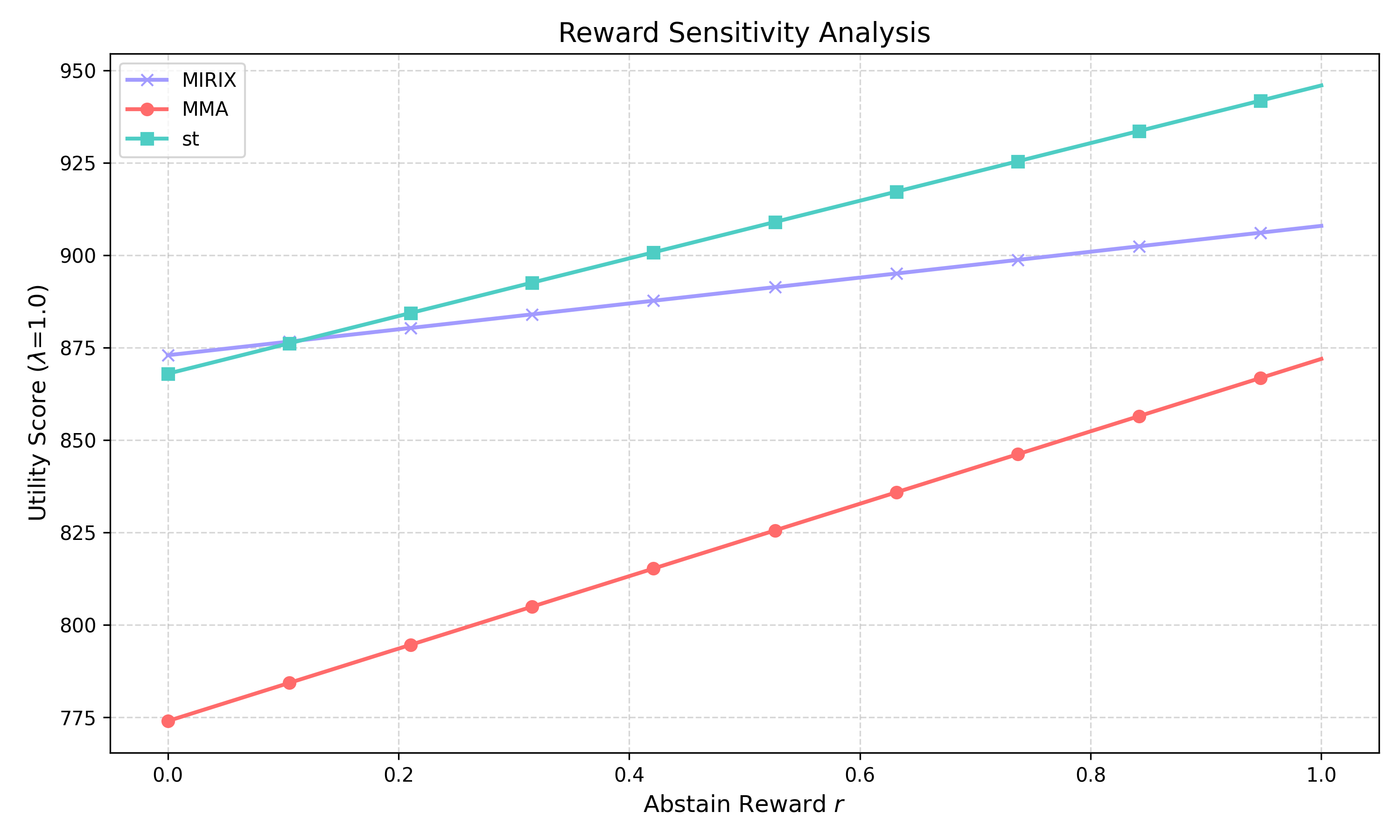}
        \caption{\textbf{Reward Sensitivity ($r$).}}
        \label{fig:locomo_reward}
    \end{subfigure}
    \caption{\textbf{Quantitative Analysis on LoCoMo.} While MMA focuses on prudence, its `st' configuration demonstrates robust utility advantages over the Baseline in high-stakes settings.}
    \label{fig:locomo_main}
\end{figure*}

We further evaluated our framework on the LoCoMo benchmark, which represents a distinct challenge: long-term conversational history with sparse information density and low adversarial conflict. We compare our Full Model (MMA) against the Baseline (MIRIX) across various reasoning dimensions. The comprehensive results are detailed in Table~\ref{tab:locomo_main} and visualized in Figure~\ref{fig:locomo_main}.

\begin{table*}[t]
\centering
\small
\resizebox{\textwidth}{!}{%
\begin{tabular}{lcccccccc}
\toprule
\multirow{2}{*}{\textbf{Method}} & \multicolumn{4}{c}{\textbf{Reasoning Categories (LLM Score)}} & \multicolumn{2}{c}{\textbf{Overall Metrics}} & \textbf{Reliability} & \textbf{Utility} \\
\cmidrule(lr){2-5} \cmidrule(lr){6-7} \cmidrule(lr){8-8} \cmidrule(lr){9-9}
& \textbf{Single-Hop} & \textbf{Multi-Hop} & \textbf{Open-Domain} & \textbf{Temporal} & \textbf{Accuracy} & \textbf{Wrong Ans.} & \textbf{Act. Acc.} & \textbf{($\lambda=1, r=0.2$)} \\
\midrule
MIRIX (Baseline) & \textbf{80.14} & \textbf{76.01} & \textbf{67.71} & 78.00 & \textbf{77.37} & 317 & 78.96\% & 880.0 \\
\textbf{MMA (Ours)} & 73.76 & 62.31 & 59.38 & 77.05 & 72.31 & 335 & 76.80\% & 793.6 \\
\midrule
\textit{Variant `st'} & 79.08 & 67.91 & 61.46 & \textbf{79.55} & 75.94 & \textbf{298} & \textbf{79.64\%} & \textbf{883.6} \\
\bottomrule
\end{tabular}%
}
\caption{\textbf{Main Results on LoCoMo.} Breakdowns of LLM Scores across reasoning dimensions ($N=1542$). While the Baseline excels in raw accuracy, our `st' variant achieves the highest \textbf{Actionable Accuracy (79.64\%)} and \textbf{Utility}, demonstrating superior reliability in safety-critical retrieval tasks.}
\label{tab:locomo_main}
\end{table*}

\paragraph{\textbf{Performance Overview.}}
As shown in Table~\ref{tab:locomo_main}, the Baseline (MIRIX) achieves a higher Overall Accuracy ($77.37\%$) and Utility Score ($573.5$) compared to MMA ($72.31\%$ / $488.0$). This performance gap is primarily driven by the Baseline's aggressive retrieval strategy (Coverage $97.73\%$), which is advantageous in non-adversarial settings where ``hallucinating'' an answer often hits the correct target by chance. In contrast, MMA adopts a significantly more prudent strategy, triggering nearly $3\times$ more abstentions ($98$ vs. $35$) due to its rigorous confidence filtering.

\paragraph{\textbf{Category-wise Analysis.}}
In the Temporal dimension, MMA achieves competitive performance ($77.05\%$) compared to the Baseline ($78.00\%$), validating the effectiveness of our Temporal Decay module in tracking timeline shifts. However, in Multi-Hop reasoning, MMA lags behind ($62.31\%$ vs. $76.01\%$). This suggests that the Conflict-Aware Consensus module, while robust against explicit contradictions (as seen in FEVER), may overly penalize weak but valid multi-hop links in sparse narrative contexts, leading to conservative ``misses'' rather than errors.

\paragraph{\textbf{Safety and Robustness.}}
Although MMA sacrifices some raw accuracy for prudence, its modular design offers flexibility. As shown in Figure~\ref{fig:locomo_utility}, the `st' variant (a configuration of MMA without consensus) successfully suppresses hallucinations, achieving the lowest wrong answer count and surpassing the Baseline in Utility ($609.0$). This highlights that while the full consensus mechanism is conservative, the core Source and Time components are highly effective for safety-critical long-context retrieval.

\begin{figure*}[t]
    \centering
    \begin{subfigure}[b]{0.48\textwidth}
        \centering
        \includegraphics[width=\textwidth]{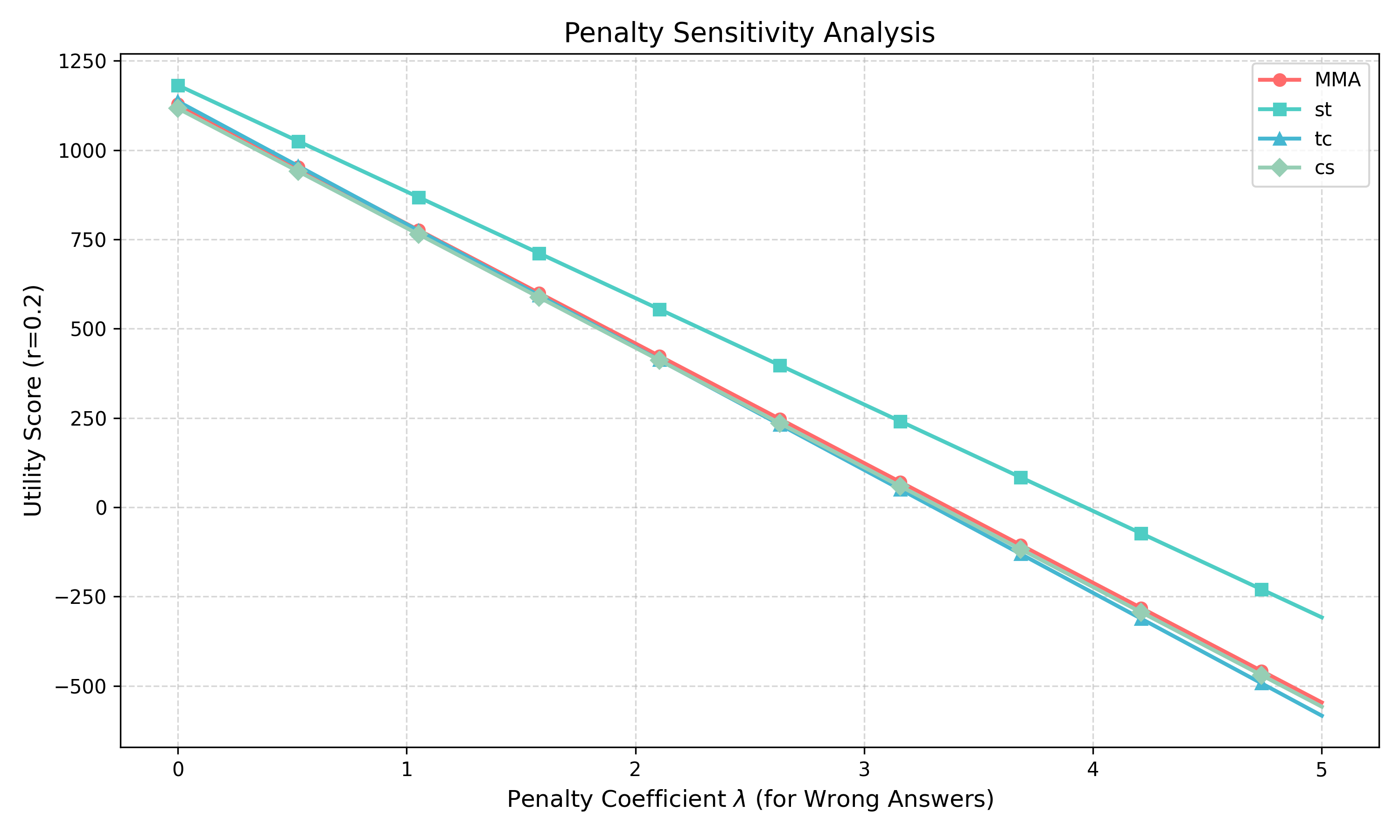} 
        \caption{\textbf{Penalty Sensitivity ($\lambda$).}}
        \label{fig:ab_penalty_locomo}
    \end{subfigure}
    \hfill
    \begin{subfigure}[b]{0.48\textwidth}
        \centering
        \includegraphics[width=\textwidth]{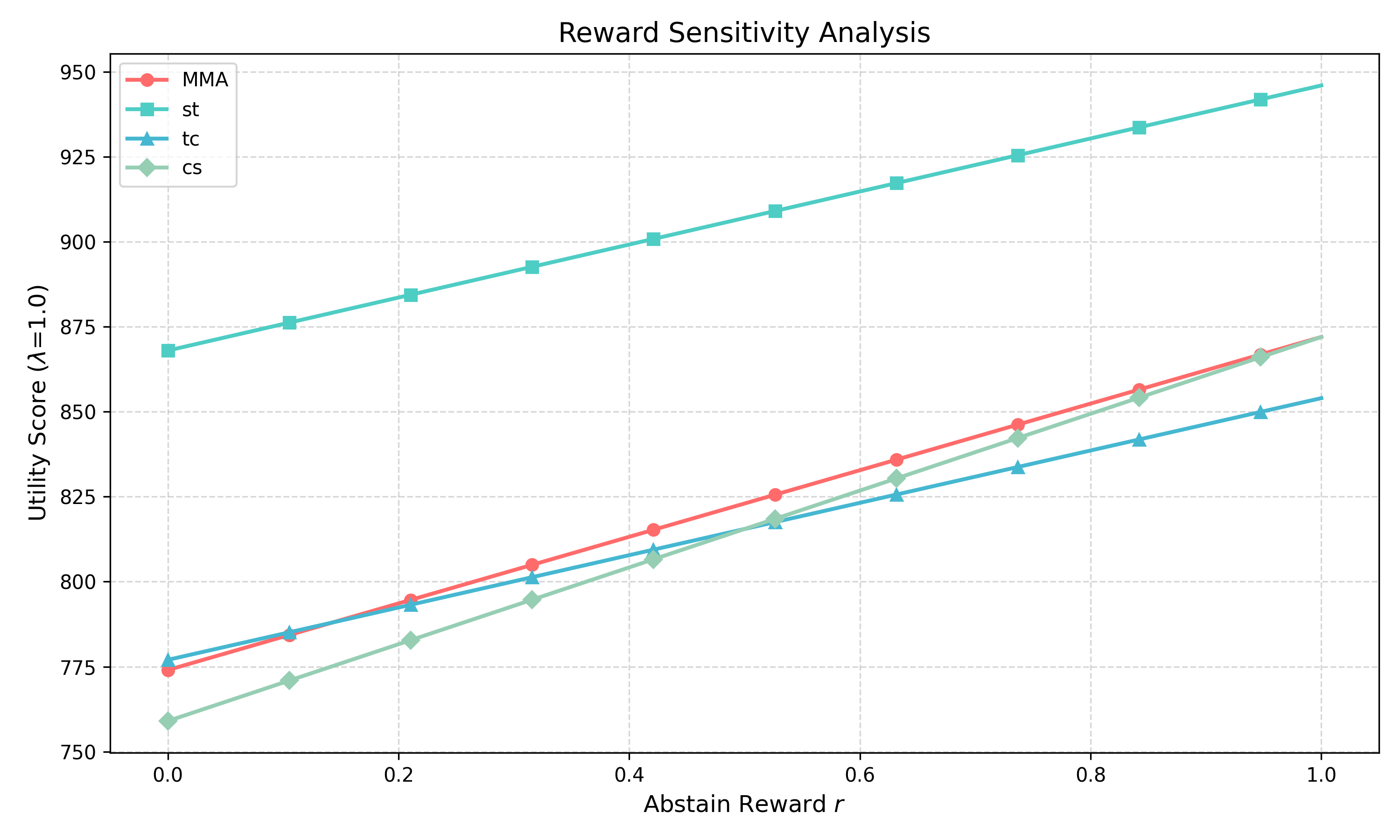}
        \caption{\textbf{Reward Sensitivity ($r$).}}
        \label{fig:ab_reward_locomo}
    \end{subfigure}
    \caption{\textbf{Ablation Sensitivity on LoCoMo.} Removing the Consensus module (`st') actually improves robustness in this specific domain, while removing Source (`tc') or Time (`cs') degrades performance.}
    \label{fig:ablation_locomo_sens}
\end{figure*}

\paragraph{\textbf{Ablation Analysis on LoCoMo.}}
Compared to the fact-centric nature of FEVER, the LoCoMo benchmark represents a distinct challenge: long-term conversational history with sparse information density. We evaluate how the removal of specific confidence components affects agent behavior in this non-adversarial but noise-heavy environment. The ablation results are summarized in Table~\ref{tab:ablation_locomo_data} and the sensitivity trends are shown in Figure~\ref{fig:ablation_locomo_sens}.

\begin{table}[h]
\centering
\small
\resizebox{\columnwidth}{!}{%
\begin{tabular}{lccccc}
\toprule
\textbf{Mode} & \textbf{Components} & \textbf{Utility} & \textbf{Wrong Ans.} & \textbf{Abstain Count} & \textbf{Act. Acc.} \\
\midrule
\textbf{MMA (Full)} & $S + T + C_{\text{con}}$ & 488.0 & 335 & 98 & 76.80\% \\
\midrule
\textbf{st} (w/o Consen.) & $S + T$ & \textbf{609.0} & \textbf{298} & 78 & \textbf{79.64\%} \\
\textbf{cs} (w/o Time) & $S + C_{\text{con}}$ & 480.5 & 335 & \textbf{113} & 76.56\% \\
\textbf{tc} (w/o Source) & $T + C_{\text{con}}$ & 471.5 & 344 & 77 & 76.52\% \\
\bottomrule
\end{tabular}%
}
\caption{Ablation results on LoCoMo ($N=1542$). \textbf{Utility} is computed with $\lambda=2.0, r=0.5$. \textbf{Wrong Ans.} denotes hallucinations (Lower is Better). The `st' variant achieves the best safety profile.}
\label{tab:ablation_locomo_data}
\end{table}

\textbf{Impact of Network Consensus ($C_{\text{con}}$):}
Contrasting with the FEVER results, removing the consensus module (Mode `st') significantly improves performance on LoCoMo, achieving the highest Utility ($609.0$) and the lowest hallucination rate ($298$ Wrong Answers). We attribute this to the ``Sparsity Paradox'': in long-term chit-chat, semantic neighbors retrieved by RAG are often thematically related (e.g., discussing dinner) but factually irrelevant to the specific query. Including these neighbors in a consensus calculation introduces noise rather than signal, diluting the confidence of correct retrievals. Thus, for sparse, non-adversarial tasks, a streamlined $S+T$ architecture is more effective.

\textbf{Impact of Source Reliability ($S$):}
The removal of the Source module (Mode `tc') results in the highest number of Wrong Answers ($344$) and the lowest Utility ($471.5$). This underscores the critical role of $S(M_i)$. In multi-turn dialogues with fixed personas, identifying and trusting reliable speakers is a primary mechanism for filtering out noise. Without this prior, the agent becomes vulnerable to misleading context, increasing the risk of hallucination. This finding validates our hypothesis that source credibility acts as a critical filter in persona-driven dialogues.

\textbf{Impact of Temporal Decay ($T$):}
The variant without time decay (Mode `cs') exhibits the highest number of Abstentions ($113$) but fails to translate this prudence into higher utility. Without the temporal dimension, the agent cannot distinguish between outdated facts and current truths, leading to a state of ``confused conservatism''—abstaining because it perceives valid updates as contradictions. This confirms that Time is essential for resolving longitudinal inconsistencies.

The ablation study reveals that while the Full MMA model is optimal for dense, adversarial verification (FEVER), the `st' configuration is superior for sparse, long-context retrieval (LoCoMo). This demonstrates the adaptability of our framework: the components can be reconfigured to match the information density of the target domain.

\section{Results on MMA-bench}
\label{sec:result_mma_bench}

\subsection{Analysis of Foundation Models}
\label{sec:analysis_models}

We evaluated two representative models, GPT-4.1-mini and Qwen3-VL-Plus, on MMA-Bench. These models were granted full context access (processing the entire dialog history at once) to isolate their reasoning capabilities from retrieval limitations. Despite this advantage, our multi-dimensional probes reveal significant deficits in their belief dynamics.

\paragraph{\textbf{Gap Between Perception and Arbitration.}}
As indicated in the breakdown of core metrics, both models demonstrate strong fundamental capabilities, achieving strong performance in fact retrieval and adversarial distraction tasks. This suggests that they effectively comprehend the long-context narrative and filter out irrelevant noise (Phase 2). However, their performance drops significantly in the 3-step probe, particularly in the verdict accuracy of conflict scenarios (ranging from 63\% to 78\%).

This discrepancy highlights a critical cognitive gap: while the models possess sufficient perception to identify the details, they lack the epistemic arbitration capability to resolve conflicts between reliable priors and contradictory visual evidence. They effectively ``read'' the text but fail to ``judge'' the truth.

\begin{table*}[t]
\centering
\small
\resizebox{\textwidth}{!}{%
\begin{tabular}{llccccc}
\toprule
\multirow{2}{*}{\textbf{Model}} & \multirow{2}{*}{\textbf{Mode}} & \multicolumn{3}{c}{\textbf{Overall Metrics}} & \multicolumn{2}{c}{\textbf{Scenario-Specific Analysis}} \\
\cmidrule(lr){3-5} \cmidrule(lr){6-7}
 & & \textbf{Core Acc.} & \textbf{Verdict Acc.} & \textbf{CoRe Score} & \textbf{Type B Acc.} & \textbf{Type D Score} \\
\midrule
\multirow{2}{*}{GPT-4.1-mini} & Text (Oracle) & 85.26\% & \textbf{77.78\%} & \textbf{0.59} & 76.47\% & \textbf{0.85} \\
 & Vision (Raw) & 80.74\% & 73.33\% & 0.51 & 64.71\% & 0.23 \\
\midrule
\multirow{2}{*}{Qwen-3-VL-Plus} & Text (Oracle) & 88.05\% & 65.56\% & 0.32 & \textbf{88.24\%} & -0.69 \\
 & Vision (Raw) & \textbf{88.98\%} & 63.33\% & 0.28 & 82.35\% & -0.69 \\
\bottomrule
\end{tabular}%
}
\caption{\textbf{Cognitive dynamics of foundation models on MMA-Bench.} \textbf{Core Acc.} measures basic reading comprehension. \textbf{CoRe Score} (Risk-Adjusted) reflects epistemic calibration. \textbf{Type B Acc.} indicates success in Reliability Inversion (overcoming authority bias). \textbf{Type D Score} reflects prudence in unknowable scenarios. Note the significant drop in Type D score for GPT-4.1-mini when switching to Vision mode, illustrating the \textit{Visual Placebo Effect}.}
\label{tab:model_performance}
\end{table*}

\begin{figure*}[t]
    \centering
    \begin{subfigure}[b]{0.48\textwidth}
        \centering
        \includegraphics[width=\textwidth]{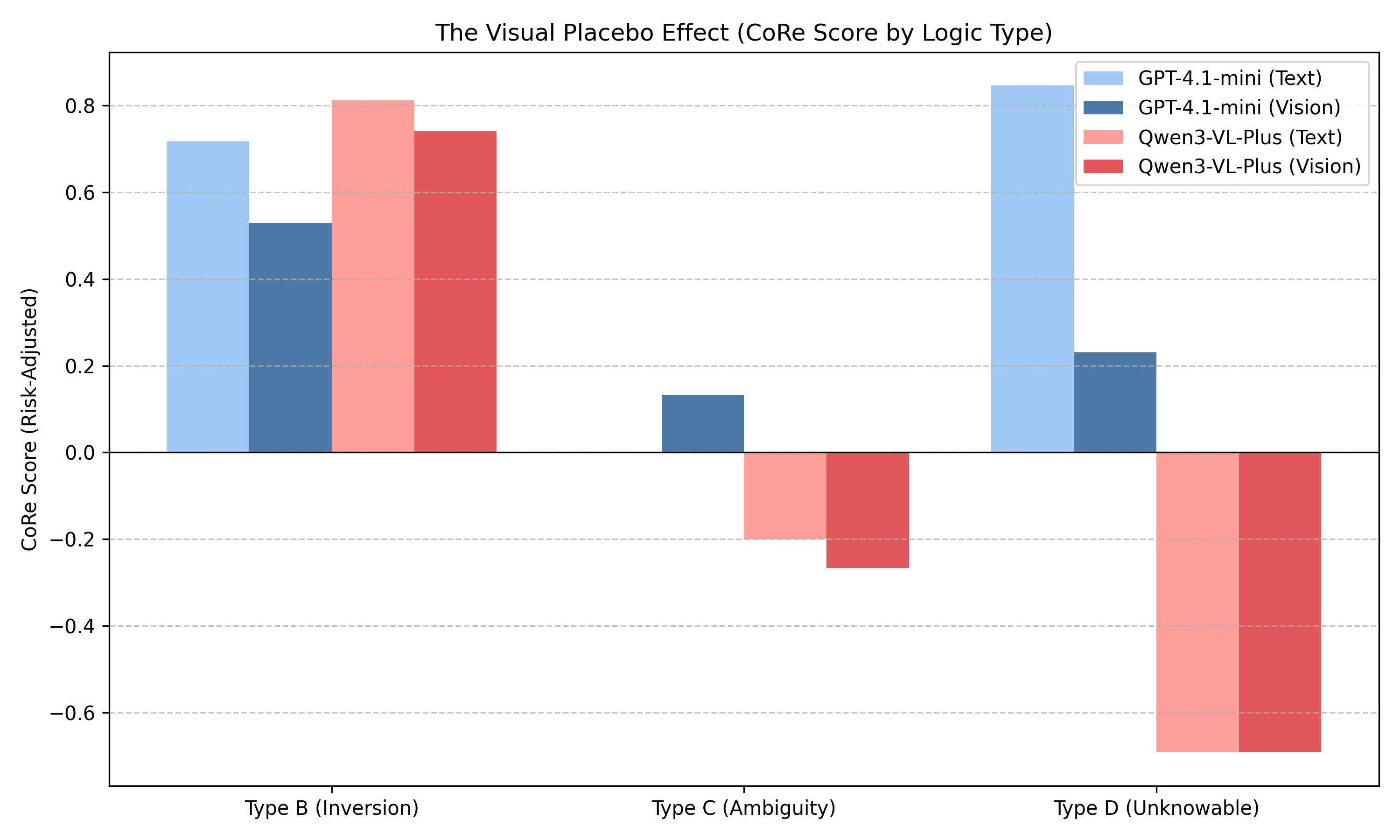} 
        \caption{\textbf{The Visual Placebo Effect.}}
        \label{fig:visual_placebo_model}
    \end{subfigure}
    \hfill
    \begin{subfigure}[b]{0.48\textwidth}
        \centering
        \includegraphics[width=\textwidth]{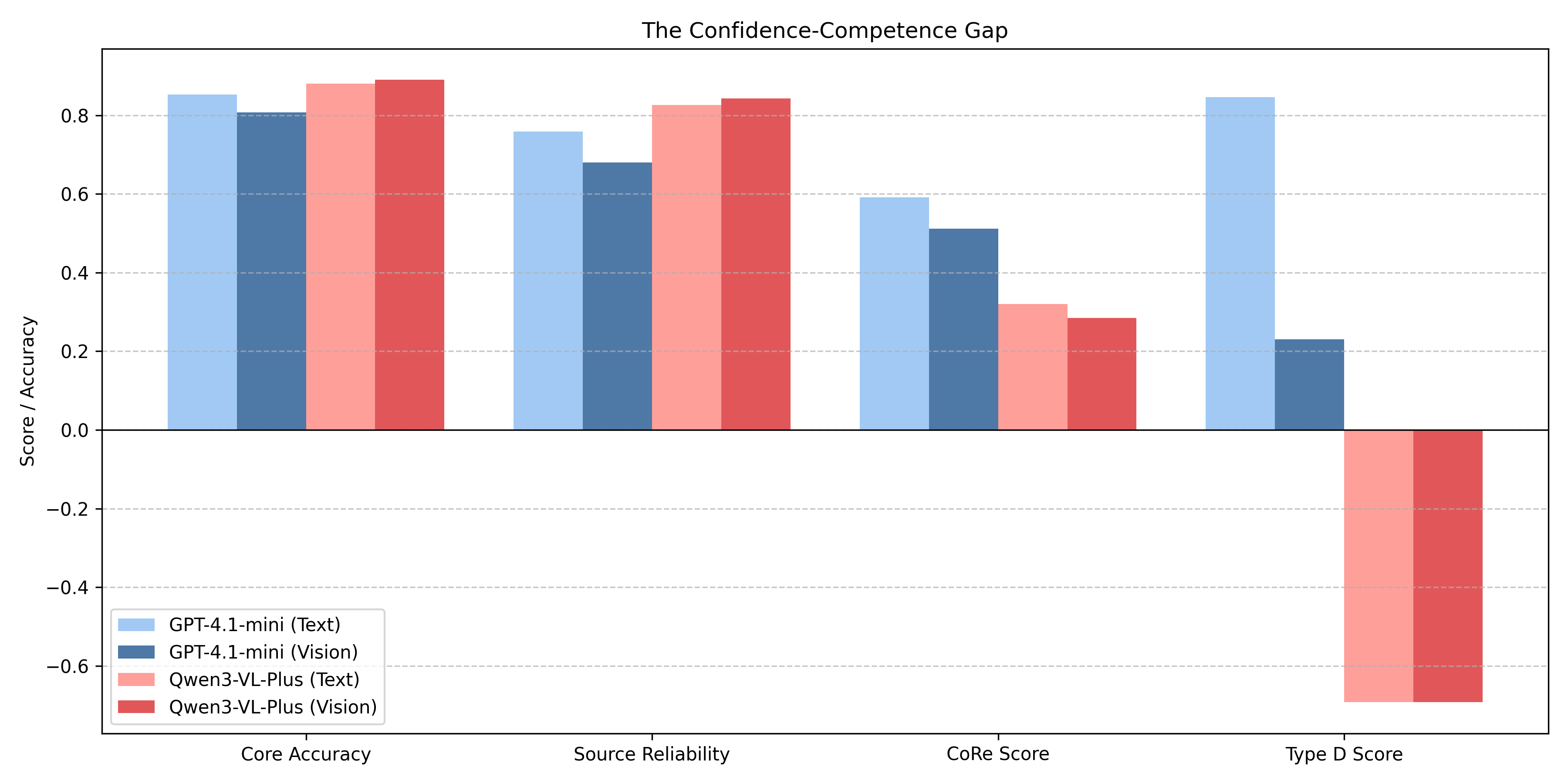}
        \caption{\textbf{The Confidence-Competence Gap.}}
        \label{fig:competence_gap_model}
    \end{subfigure}
    
    \caption{\textbf{Cognitive Dynamics of Foundation Models on MMA-Bench.} We compare GPT-4.1-mini and Qwen-3-VL-Plus across Text (Oracle) and Vision (Raw) modes. (a) Reveals how visual modalities can act as distractors in noise scenarios. (b) Highlights the disconnect between reading comprehension (Core Acc) and epistemic prudence (CoRe Score).}
    \label{fig:model_analysis_model}
\end{figure*}

\paragraph{\textbf{Modality Preference and Visual Placebo Effect.}}
We utilized the modality signal alignment metric to diagnose how visual inputs influence decision-making. The results expose divergent behaviors between the two models.

Type B (Inversion) scenarios reveal strong authority bias. Both models struggle to consistently prioritize objective visual evidence over textual statements from a historically reliable source (User A). Qwen3-VL-Plus exhibits a stronger tendency towards visual grounding (82.4\% alignment with visual signals) compared to GPT-4.1-mini (64.7\%), reflecting its architectural strength in vision. However, a significant portion of errors stems from the models hallucinating a justification to align the visual evidence with the textual prior.

In indeterminate scenarios (Type C and D), we observe a phenomenon we term the visual placebo effect. For GPT-4.1-mini, performance in Type D (Unknowable) scenarios degrades drastically when switching from text mode (oracle captions) to vision mode (raw images), with the CoRe score dropping from 0.85 to 0.23. This suggests that the presence of an image, even if irrelevant or ambiguous, creates an illusion of information sufficiency, prompting the model to fabricate definitive answers rather than maintain prudence. Conversely, Qwen3-VL-Plus exhibits extreme overconfidence in these noise scenarios across both modes, frequently placing high wagers on hallucinated verdicts, indicating a fundamental lack of epistemic calibration.

\paragraph{\textbf{Fragility of Self-Correction.}}
Our analysis of the confession mechanism (Step 3) reveals pathological instability in reasoning. Although both models achieve high self-correction rates numerically, qualitative inspection shows that over 50 cases involved the models flipping from a correct verdict to an incorrect one during the reflection phase.

This behavior suggests that the self-correction mechanism is impelled not by authentic introspection but by instructional sycophancy, a propensity whereby the model conforms to the skepticism implicitly encoded in the reflection prompt. Furthermore, we observed a prevalence of logic collapse, where models would place a high wager on a verdict in Step 2, only to immediately confess it was wrong in Step 3. This disconnect between the acting system (wagering) and the thinking system (reflecting) underscores the immaturity of current models in maintaining a coherent belief state.

\subsection{Ablation on MMA-Bench}

\begin{figure*}[h!]
    \centering
    \begin{subfigure}[b]{0.48\textwidth}
        \centering
        \includegraphics[width=\textwidth]{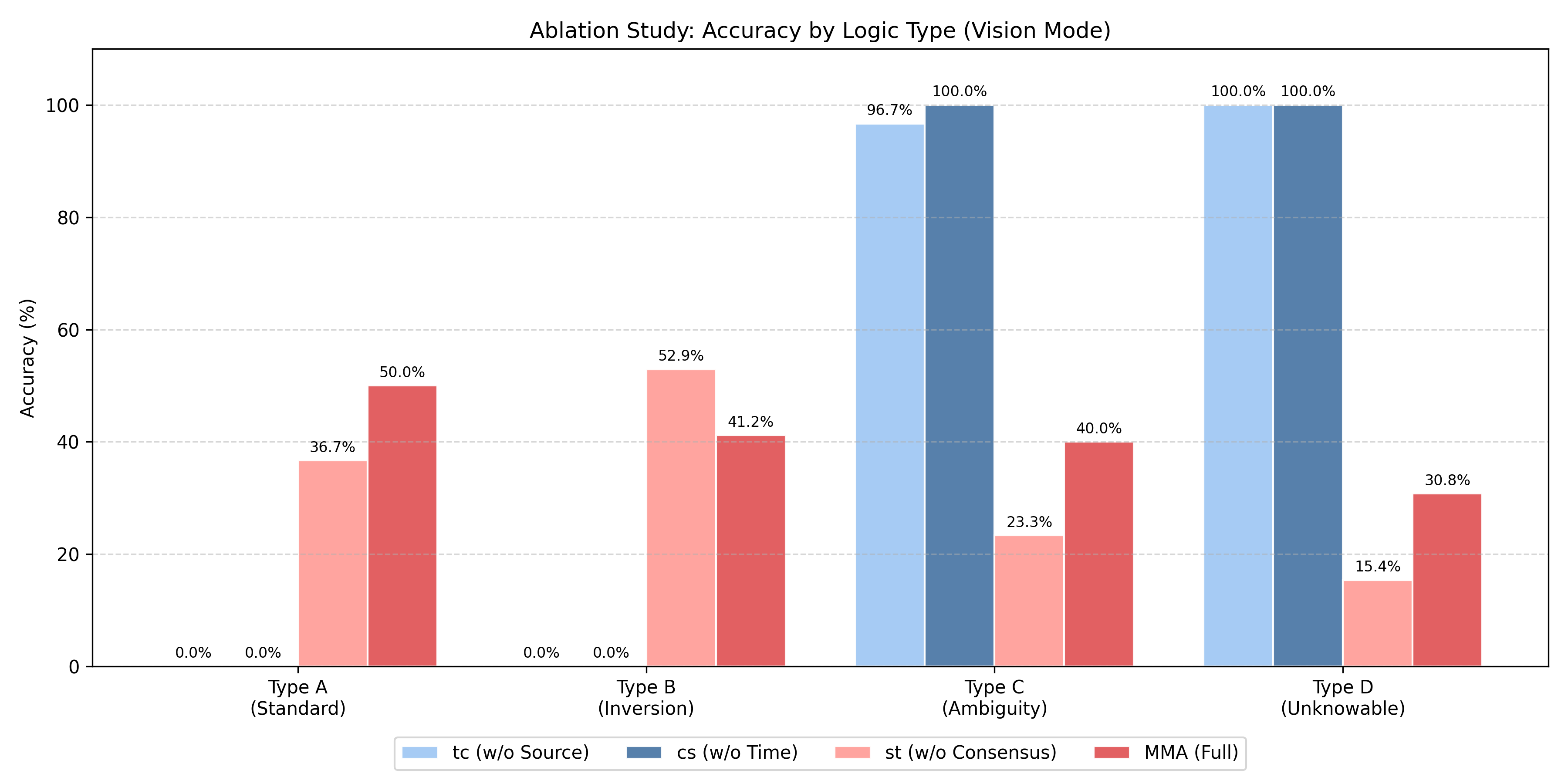} 
        \caption{\textbf{Cognitive Paralysis (Accuracy Analysis).}}
        \label{fig:ablation_paralysis_mma}
    \end{subfigure}
    \hfill
    \begin{subfigure}[b]{0.48\textwidth}
        \centering
        \includegraphics[width=\textwidth]{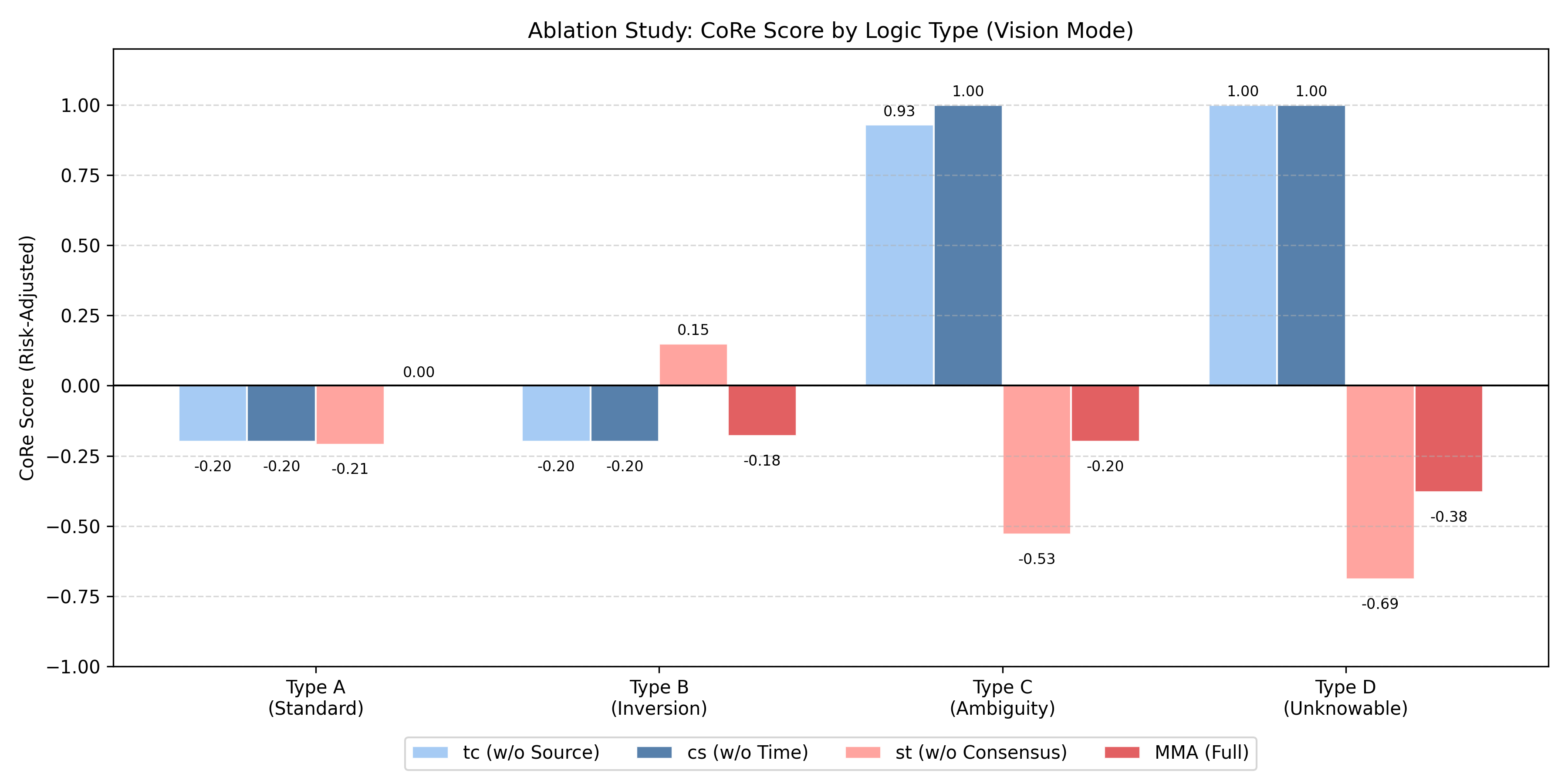}
        \caption{\textbf{Visual Placebo Mitigation (CoRe Score Analysis).}}
        \label{fig:ablation_placebo_mma}
    \end{subfigure}
    \caption{\textbf{Mechanism Ablation on MMA-Bench (Vision Mode).} We isolate the failure modes: (a) \textbf{Accuracy metrics} reveal that Source ($S$) and Time ($T$) are prerequisites for agency, as their absence leads to paralysis (0\% accuracy in known facts); (b) \textbf{CoRe Scores} demonstrate that Consensus ($C_{\text{con}}$) is essential to buffer against the \textbf{Visual Placebo Effect} in indeterminate queries.}
    \label{fig:ablation_main_mma}
\end{figure*}

To dissect the specific mechanisms driving the cognitive behaviors observed in Subsection \ref{subsec:analysis_agent}, we evaluated three ablated variants against the Full Model ($S+T+C_{\text{con}}$) on MMA-Bench. The results, summarized in Table \ref{tab:ablation_mma_data} and visualized in Figure \ref{fig:ablation_main_mma}, isolate the distinct contributions of each component.

\paragraph{\textbf{Impact of Source Reliability ($S$):}}
Comparison with Mode `tc' (w/o Source) reveals that source credibility is a prerequisite for agency. Without the source module, the agent exhibits symptoms of cognitive paralysis. We demonstrate this by contrasting performance across logic types: while Mode `tc' achieves superficially perfect scores in indeterminate scenarios (Type D: $1.0$, Type C: $96.7\%$), it paradoxically yields 0.0\% accuracy in all deterministic scenarios (Type A and Type B) (Table \ref{tab:ablation_mma_data}). This distinct data pattern, visualized in Figure \ref{fig:ablation_paralysis_mma}, indicates that the agent is not exercising prudence but is mechanically incapable of forming positive verdicts. Lacking a prior trust distribution, it defaults to ``Unknown'' for every query. Thus, unlike MMA which demonstrates functional discrimination (Vision Type A: $50.0\%$), the success of `tc' in indeterminate cases is merely a statistical artifact of system inaction.

\paragraph{\textbf{Impact of Network Consensus ($C_{\text{con}}$):}}
Mode `st' (w/o Consensus) highlights the role of consensus in mitigating the visual placebo effect. The results reveal an intriguing trade-off: without the consensus constraint, `st' is more aggressive in accepting visual evidence, actually outperforming MMA in Type B Inversion scenarios ($52.9\%$ vs. $41.2\%$). However, this aggression proves fatal in indeterminate contexts. In Vision Mode, its Type D score collapses catastrophically to $-0.69$, indicating that isolated visual signals override textual caution (Figure \ref{fig:ablation_placebo_mma}). In contrast, the Full Model employs $C_{\text{con}}$ to validate visual inputs against the semantic neighborhood. While this conservatism slightly dampens Type B performance, it significantly buffers the Type D drop (Score: $-0.38$), providing a critical safety layer against hallucination.

\paragraph{\textbf{Impact of Temporal Decay ($T$):}}
Mode `cs' (w/o Time) demonstrates a critical failure in stability when shifting modalities. We observe that while `cs' performs comparably to MMA in Text Mode (Type A Acc: $40.0\%$), its capability degrades significantly in Vision Mode, dropping to 0.0\% in Type A scenarios (see Table \ref{tab:ablation_mma_data}). This distinct drop suggests that, once temporal decay is removed, the historical noise that remains tolerable within the confines of textual input accumulates without bound; when such accumulated noise is further compounded by high-dimensional visual features, the signal-to-noise ratio is driven below the decision threshold. MMA utilizing $T$ maintains consistent performance across modes (Vision Type A: $\sim 50\%$), proving that temporal awareness is essential for robustness in high-entropy multimodal environments.

\end{document}